\documentclass[sigconf]{acmart}
\usepackage{microtype}
\usepackage{graphicx}
\usepackage{tikz}
\usepackage{makecell}
\usepackage{bussproofs}
\usepackage{pifont}
\usepackage{hyperref}
\usepackage{xspace}
\usepackage{algorithm}
\usepackage{algorithmic}
\usepackage{multirow}
\usepackage{booktabs}

\newif\ifdraft
\drafttrue 

\newcommand{\xg}{XGrammar-2\xspace}

\newcommand{\secref}[1]{\S\ref{#1}}

\renewcommand{\algorithmiccomment}[1]{\{#1\}}

\makeatletter
\newcommand*\bigcdot{\mathpalette\bigcdot@{.7}}
\newcommand*\bigcdot@[2]{\mathbin{\vcenter{\hbox{\scalebox{#2}{$\m@th#1\bullet$}}}}}
\makeatother

\AtBeginDocument{%
  }

\setcopyright{acmlicensed}
\copyrightyear{2026}
\acmYear{2026}
\acmDOI{XXXXXXX.XXXXXXX}

\begin{document}

\title{\xg: Dynamic and Efficient Structured Generation
Engine for Agentic LLMs}

\author{Linzhang Li}
\authornote{Both authors contributed equally to this research.}
\email{blemiade_qinchuan@sjtu.edu.cn}
\affiliation{
  \institution{Shanghai Jiao Tong University}
  \country{China}
}

\author{Yixin Dong}
\authornotemark[1]
\authornote{Corresponding authors.}
\email{yixind@andrew.cmu.edu}
\affiliation{
  \institution{Carnegie Mellon University}
  \country{USA}
}

\author{Guanjie Wang}
\email{irfnfnkemed@sjtu.edu.cn}
\affiliation{
  \institution{Shanghai Jiao Tong University}
  \country{China}
}

\author{Ziyi Xu}
\email{xzy2022@sjtu.edu.cn}
\affiliation{
  \institution{Shanghai Jiao Tong University}
  \country{China} 
}

\author{Alexander Jiang}
\email{akj2@andrew.cmu.edu}
\affiliation{
  \institution{Carnegie Mellon University}
  \country{USA}
}

\author{Tianqi Chen}
\authornotemark[2]
\email{tqchen@cmu.edu}
\affiliation{
  \institution{Carnegie Mellon University, NVIDIA}
  \country{USA}
}



\acmYear{2026}\copyrightyear{2026}
\setcopyright{cc}
\setcctype[4.0]{by}
\acmConference[ACM CAIS '26]{ACM Conference on AI and Agentic Systems}{May 26--29, 2026}{San Jose, CA, USA}
\acmBooktitle{ACM Conference on AI and Agentic Systems (ACM CAIS '26), May 26--29, 2026, San Jose, CA, USA}
\acmDOI{10.1145/3786335.3813124}
\acmISBN{979-8-4007-2415-2/26/05}

\begin{abstract}
    Modern LLM agents increasingly rely on dynamic structured generation, such as tool calling and response protocols. Unlike traditional structured generation with static structures, these workloads vary both across requests and within a request, posing new challenges to existing engines. We present \xg, a structured generation engine for dynamic agentic workloads. Our design is based on two key ideas: first-class support for tag-triggered structure switching, and fine-grained reuse across requests with different output structures. Concretely, \xg introduces TagDispatch for dynamic structural dispatching and Cross-Grammar Cache for substructure-level cache reuse across grammars. It further improves efficiency with an Earley-based adaptive token mask cache, just-in-time compilation, and repetition state compression. Experiments show that \xg achieves over 6$\times$ faster compilation than prior structured generation engines, and incurs near-zero end-to-end overhead in modern LLM serving systems.

\end{abstract}


\maketitle

\section{Introduction}

Modern LLM agents demonstrate strong capabilities and increasingly rely on complex tool calling and code generation~\cite{Park2023GenerativeAgents}. These agentic applications impose strong requirements on structured generation, especially for small~\cite{patil2025bfcl} or compressed models. Constrained decoding~\cite{deutsch-etal-2019-general, kuchnik2023validating} is widely adopted to guarantee structural validity by masking invalid tokens at each generation step, enabling reliable downstream applications with minimal overhead.

\begin{figure}[h]
    \centering
    \includegraphics[width=\linewidth]{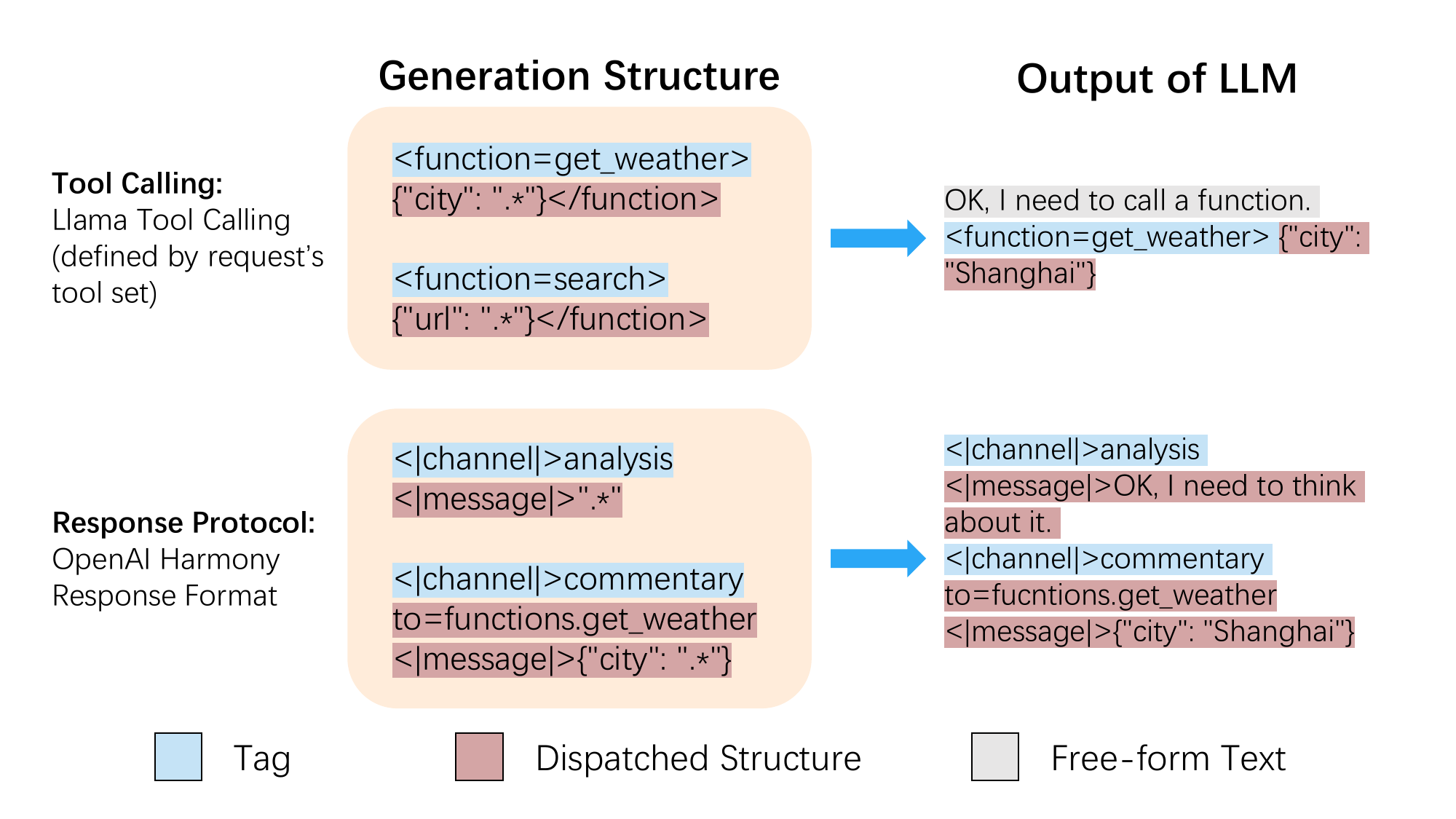}
    \caption{Some examples of tool calling and response protocols.}
    \label{fig: examples}
\end{figure}

\begin{figure*}[t]
    \centering
    \includegraphics[width=\linewidth]{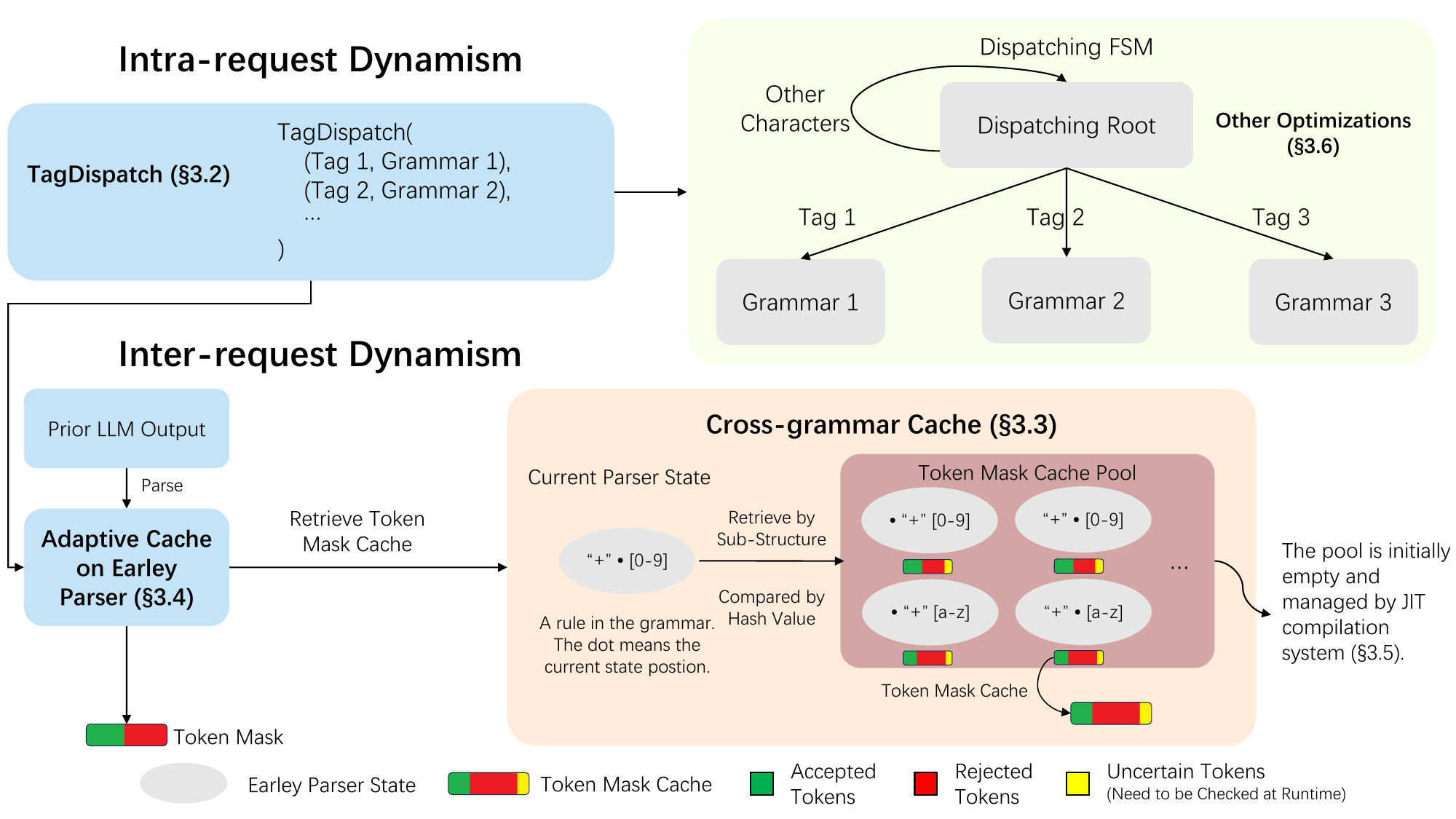}
    \caption{Overview of our approach. We design a new dynamic dispatching semantics, TagDispatch (\secref{sec:tagdispatch}), to efficiently support intra-request dynamism. To leverage the sub-structures across different grammars, we designed a cross-grammar caching algorithm (\secref{cgc}) based on the Earley parser (\secref{earley}) to handle inter-request dynamism. We also design a JIT compilation method (\secref{jit}) to optimize the efficiency for the inter-request dynamism. We also introduce a repetition state compression algorithm (\secref{repetition}) to handle repetition structures.}
    \label{fig: overview}
\end{figure*}

However, existing constrained decoding methods~\cite{xgr, githubGitHubllguidance, willard2023efficientguidedgenerationlarge} largely assume all structures are static and known in advance. Nowadays, a key characteristic of agentic LLM applications is the extensive use of tool calling to handle complex tasks. Each LLM request may contain dozens or even hundreds of possible tools, which greatly violates the structure assumption: the output structure becomes highly dynamic, both across requests and within a single request.  This \textbf{structural dynamism} poses significant efficiency and expressiveness challenges to existing constrained decoding systems. We classify the challenge of structural dynamism into two categories:

\paragraph{Inter-request dynamism.} In agent serving scenarios, each request may expose a different set of tools and schemas, often with per-tool access control~\cite{microsoft_foundry_agents_overview_2026, openai_apps_in_chatgpt_2025}. As a result, the space of possible output grammars becomes combinatorially large, and each grammar can itself be complex. Prior approaches typically preprocess the entire grammar and cache it at the request level to reuse identical structures. Under dynamic tool sets, such caching becomes ineffective, forcing expensive per-request preprocessing and significantly increasing time-to-first-token (TTFT).

\paragraph{Intra-request dynamism.}  Within a single request, the model needs to follow a response protocol such as OpenAI Harmony~\cite{kundel2025openaiharmony}, and choose from many candidate tools. This requires the structural constraint to switch depending on the previous LLM output. For example, generating a tool name determines the JSON schema of the subsequent arguments~\cite{meta2024toolcalling, qwen_function_calling}, while a channel tag token constrains the following content to a specific channel, such as reasoning or output. Such dispatching is difficult to express efficiently with the Backus-Naur Form (BNF)-like grammars used by existing constrained decoding methods, and the large number of tools further challenges efficient mask generation.

To address these challenges, we propose \xg, a structured generation engine for dynamic agentic workloads. Our design is based on two key ideas: first-class support for tag-triggered structure switching in agent outputs, and fine-grained reuse across requests with different output structures. For the former, we introduce TagDispatch, a first-class grammar construct for expressing tag-triggered structural dispatching within a request. For the latter, we design a cross-grammar cache that reuses shared substructures across different grammar combinations. To make this design efficient in practice, we further develop an Earley-based adaptive token mask cache, together with just-in-time compilation and repetition compression, to reduce compilation overhead and improve end-to-end efficiency.

We implement \xg as a structured generation engine compatible with modern LLM inference systems. \xg supports tool-calling formats across major models and enforces strict compliance with the OpenAI Harmony Response Format~\cite{kundel2025openaiharmony}. Experimental results show that \xg achieves over 6× tool-calling compilation speed improvement compared to prior state-of-the-art methods, while introducing near-zero latency overhead. We have incorporated \xg into open-source serving frameworks such as SGLang~\cite{zheng2024sglangefficientexecutionstructured} and vLLM~\cite{kwon2023efficient}, improving output reliability in agentic tasks. \xg is open-source and has been adopted in both industry systems and open-source inference engines.

\section{Background}

\subsection{Constrained Decoding and Context-free Grammar}

LLMs like Deepseek-R1 \cite{deepseekai2025deepseekr1incentivizingreasoningcapability}, gpt-oss \cite{openai2025gptoss120bgptoss20bmodel} all generate the tokens autoregressively, predicting the next token based on the previous output. Each time the LLM needs to output a token, it will calculate a logit vector for the vocabulary and then convert it into a probability distribution with the softmax function~\cite{NIPS1989_0336dcba}. In the end, a sampler will choose an output token based on the distribution to output.

Constrained decoding \cite{deutsch-etal-2019-general} is a technique for guiding LLMs to generate text according to a specified grammar. During each decoding step, tokens that do not conform to the grammar are marked as invalid, and their corresponding logit values are set to $-\infty$ to assign them zero probability, thus preventing them from being sampled and ensuring the output of LLMs follows the grammar. 

Context-free Grammar (CFG)~\cite{1056813} is generally used to define the grammar structures, and it is described by Extended Backus-Naur Form (EBNF)~\cite{iso14977} in most constrained decoding methods. An EBNF consists of a set of production rules, each representing a symbol that can be expanded into a sequence of terminal characters or references to other symbols. With the rule references, EBNF can naturally express complex recursive structures.

\subsection{XGrammar}
 
Constrained decoding modifies the logit vector before the LLM outputs the next token, requiring a runtime check to determine whether the token is valid across the entire vocabulary. Without optimization, this process introduces significant overhead, which substantially slows down the output speed of LLMs.

XGrammar \cite{xgr} is designed to achieve near-zero overhead token mask generation. XGrammar employs a pushdown automaton parser to trace the output of LLMs. Its key insight is that for each state in CFGs, there are a lot of tokens that can be determined to be accepted or rejected within the state's rule, and there are a few context-dependent tokens that need the context information to determine whether they can be accepted by the current state at runtime. XGrammar stores the pre-computed accepted tokens, rejected tokens, and context-dependent tokens into the adaptive token mask cache. With the token mask cache, XGrammar can skip massive computation for accepted tokens and rejected tokens at runtime. Moreover, XGrammar further increases the cache hit rate by introducing context expansion, which leverages the rule reference structure in the grammar to further check and reject context-dependent tokens.

With the optimization techniques, XGrammar can handle static structured generation tasks well. However, XGrammar needs to compile all the grammars ahead of time, which is not suitable for dynamic structured generation tasks, since the grammars can be sent to the engine at runtime. Thus, how to efficiently handle dynamic structured generation tasks remains a challenge.

\section{Methods}
\subsection{Overview}


\xg addresses dynamic agentic workloads with a unified design centered on first-class structural dispatching and fine-grained reuse across dynamically changing grammars. TagDispatch (\autoref{sec:tagdispatch}) captures intra-request dynamism by expressing tag-triggered switching between free-form text and structured sub-grammars. Cross-grammar cache (\autoref{cgc}) handles inter-request dynamism by reusing token mask caches across grammars with shared substructures. To support efficient execution on dynamic and complex grammars, \xg adopts an Earley-based adaptive token mask cache (\autoref{earley}) as the cache mechanism. JIT compilation (\autoref{jit}) further amortizes cache construction over decoding steps instead of materializing the full cache upfront. Repetition state compression (\autoref{repetition}) reduces runtime overhead and improves robustness for recurring grammar patterns.


\subsection{TagDispatch: Dynamic Dispatch Semantics}
\label{sec:tagdispatch}

Intra-request dynamism: prior output determines subsequent structures.
This requires free-formed text interleaved by structure constraints separated by certain triggers, such as a tool name or a channel control token. Although this semantics can in principle be encoded in plain EBNF, the encoding becomes cumbersome and inefficient, since it must simultaneously accept arbitrary non-tag text, recognize multiple tags, and route each tag to a different sub-grammar.

\begin{figure}[htbp]
    \centering
    \includegraphics[width=\linewidth]{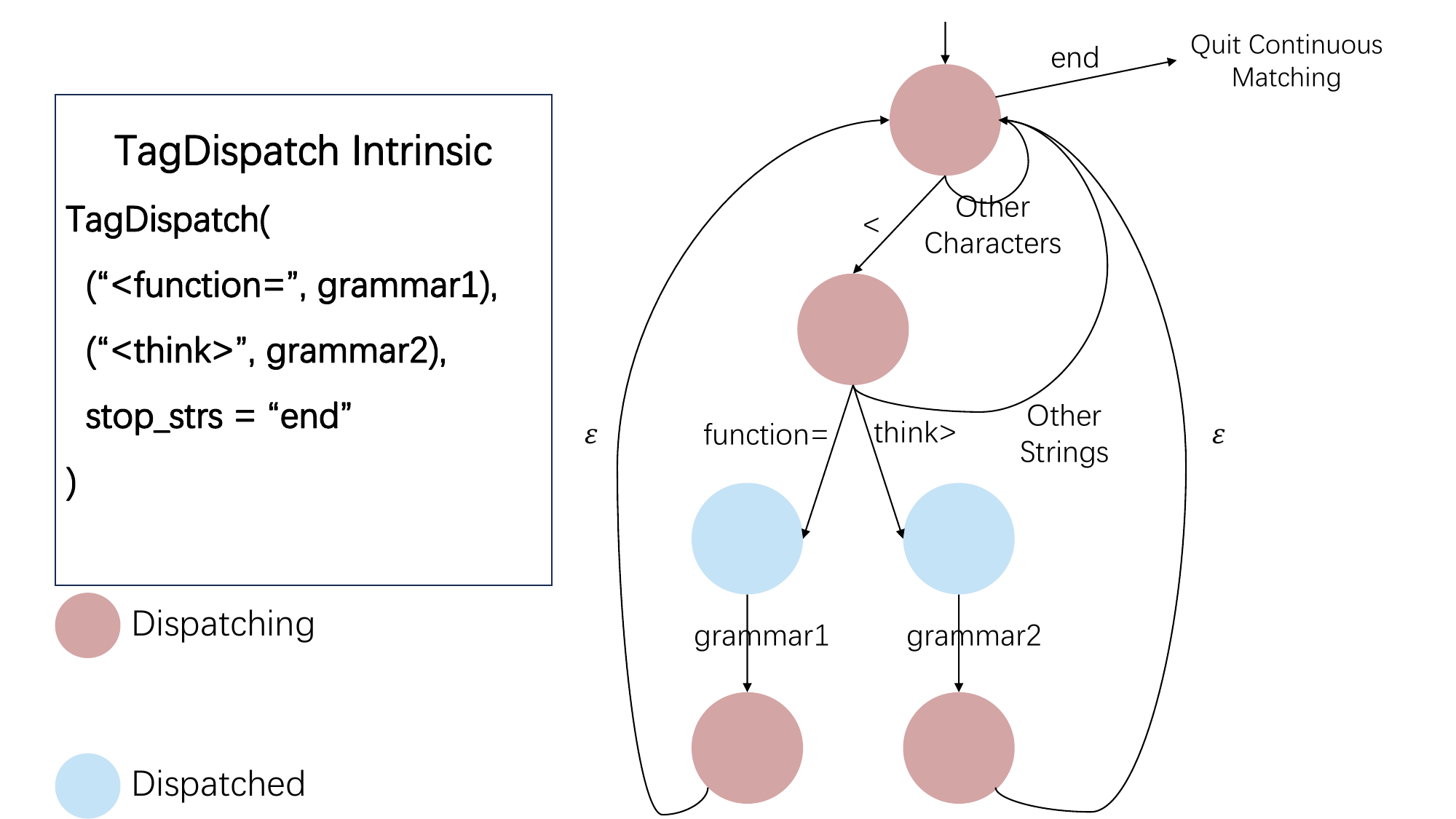}
    \caption{The definition and the constructed automata from TagDispatch.}
    \label{fig:tagdispatch}
\end{figure}

To effectively express such structures, we introduce TagDispatch, an EBNF-compatible grammar intrinsic to describing tag-triggered switching between free-form text and structured sub-grammars. As shown in~\autoref{fig:tagdispatch}, a \texttt{TagDispatch} is parameterized by (i) a list of tag--grammar pairs $(t_i, G_i)$, where emitting tag $t_i$ dispatches decoding to sub-grammar $G_i$, and (ii) a set of stop strings \texttt{stop\_strs} that terminate dispatching. Conceptually, TagDispatch partitions decoding into two modes: \emph{dispatching} and \emph{dispatched}. Decoding starts in the dispatching mode, where the engine accepts ordinary text while continuously matching registered tags. Once a tag is matched, the engine switches to the dispatched mode and constrains subsequent decoding with the corresponding sub-grammar. After that sub-grammar completes, decoding returns to the dispatching mode. If a stop string is matched in the dispatching mode, TagDispatch exits.


In the dispatching mode, we use an Aho–Corasick automaton (AC automaton)~\cite{10.1145/360825.360855} to match multiple tags simultaneously. The automaton compiles all candidate tags into a single deterministic finite automaton (DFA), enabling incremental matching over the generated text. When a partial match fails, the automaton falls back to a previously matched state and continues matching. This enables efficient online trigger matching over free-form text.

TagDispatch can effectively describe agentic output structures. For example, a snippet of LLM output with tool calling is \texttt{OK, I will call a tool. <function=get\_weather>\{"city":"San Francisco"\} </function>}.
The prefix \texttt{<function=get\_weather>} can be registered as a tag in TagDispatch, and dispatches decoding to the JSON-argument grammar (and optional wrapper grammar) associated with \texttt{get\_weather}. After the dispatched grammar completes, TagDispatch returns to the dispatching mode, allowing the model to continue generating free-form text or trigger another tag. The same abstraction also applies to channelized outputs, where a channel tag is followed by a channel-specific structure.

\subsection{Cross-Grammar Cache}
\label{cgc}

Different requests' grammars often share some common sub-structures. Even within a single grammar, some sub-structures are still duplicated. These repeated compilation leads to large overhead. To leverage the token mask caches of these sub-structures, we design a \textbf{Cross-Grammar Cache} to avoid recomputation.

In \xg, structures are represented as multiple FSMs. Each FSM can have edges referring to another FSM to represent the recursive structure in EBNF. To efficiently reuse the token mask caches of the common sub-structures, we have two main challenges:

\begin{enumerate}
    \item \textit{How to detect the common substructures.} We need to determine whether two FSMs are equivalent; since each FSM can refer to other FSMs, the checker also needs to check the referred FSM, and the reference structure may contain loops.
    \item \textit{How to reuse the token mask caches from other FSMs.} In XGrammar, the token mask cache not only considers FSM's structural information, but also how this FSM is referred to by other FSMs to further increase cache hit rate (see context-expansion in XGrammar paper). Even though the structure of two FSM matches, the cache may not be simply reused because they have different referencing structure.~\cite{xgr}.
\end{enumerate}

For the first challenge, we design a \textbf{hierarchical hashing algorithm} for FSMs to detect identical sub-structures. This algorithm resolves the problem by assigning each FSM a structural hash that incorporates both its local state-transition structure and the whole-structure hashes of the FSMs referenced by its rule-reference edges. The key idea is to combine the hash of each referenced FSM into the hash of the referencing FSM, so that structural information is aggregated bottom-up along the FSM reference graph. Cyclic references break this bottom-up order and therefore require additional handling. The overall procedure is:
\begin{enumerate}
    \item Build the FSM reference graph induced by rule-reference edges.
    \item Hash the acyclic portion bottom-up with Algorithm~\autoref{alg:fsm_hash}.
    \item Handle each simple cycle using provisional hashes followed by cycle-hash refinement (Algorithm~\autoref{alg: cycle}).
    \item Use the final FSM hashes as keys for cross-grammar cache reuse.
\end{enumerate}

Algorithm~\autoref{alg:fsm_hash} hashes one FSM, assuming that the hashes of all referenced FSMs are already available. It first canonicalizes the local state graph by deterministically sorting outgoing edges and assigning canonical state IDs via BFS from the initial state. It then traverses the states in this canonical order and incrementally hashes the serialized state and edge information, including the edge type, label, and target state ID. Therefore, for the acyclic portion of the reference graph, we can topologically sort the FSMs and apply Algorithm~\autoref{alg:fsm_hash} in reverse topological order.

Simple cycles require additional handling because the bottom-up assumption of Algorithm~\autoref{alg:fsm_hash} no longer holds: an FSM in the cycle may refer to another FSM whose final hash is not yet known. To address this, we first assign a special provisional value to unresolved rule-reference edges inside the cycle and apply Algorithm~\autoref{alg:fsm_hash} to obtain provisional hashes for the FSMs in the cycle. We then apply Algorithm~\autoref{alg: cycle}~\cite{DBLP:journals/corr/abs-2002-06653} to refine these provisional hashes with the cycle structure itself. This yields distinct final hashes for different positions in the cycle and preserves the uniqueness of the resulting structural hashes.

\begin{figure}[htbp]
    \centering
    \includegraphics[width=\linewidth]{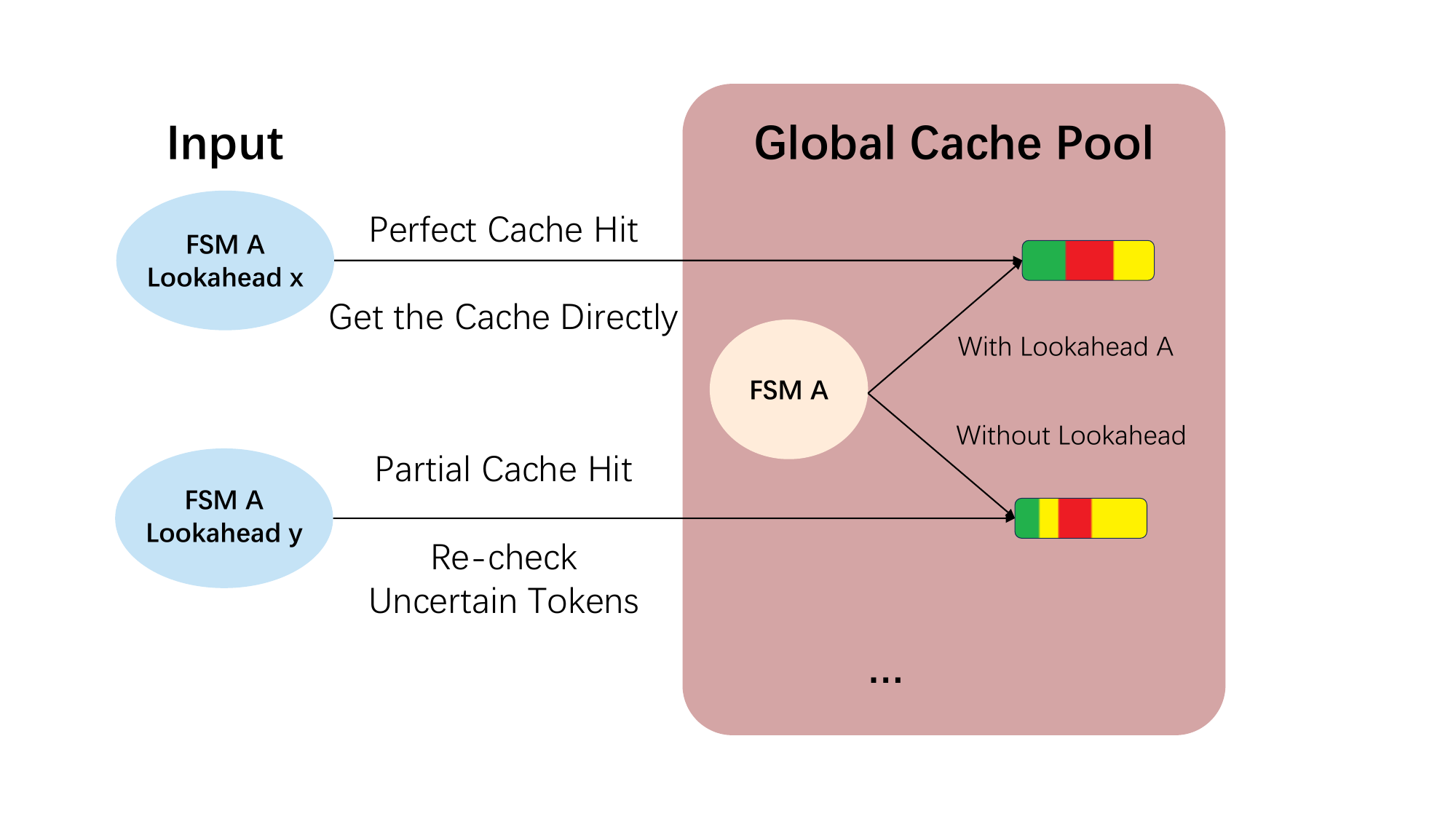}
    \caption{Cross-grammar cache reuse under matching and mismatched lookahead conditions.
    }
    \label{fig: cgc}
\end{figure}

\begin{algorithm}[t]
\caption{Canonical Hash of One FSM Given Referenced FSM Hashes}
\label{alg:fsm_hash}
\begin{algorithmic}
\STATE {\bfseries Input:} Finite state machine $\mathcal{A} = (S, E, F, s_0)$, where $S$, $E$, $F$, and $s_0$ denote the state set, edge set, final-state set, and initial state
\STATE {\bfseries Input:} For every rule-reference edge $e \in E$, the hash of the referenced FSM $h(e.\mathit{ref})$ is already available
\STATE {\bfseries Output:} Canonical structural hash $h$ of $\mathcal{A}$

\STATE {\bfseries Hash function:} Let $\mathcal{H}$ be an order-sensitive hash function over sequences
\STATE {\bfseries Constants:} $\mathit{NODE\_TAG}, \mathit{RANGE\_TAG}, \mathit{REF\_TAG}, \mathit{EPS\_TAG}$

\vspace{0.5em}
\STATE {\bfseries Phase 1: Canonical state ordering}
\STATE Sort the outgoing edges of each state in the following order:
\STATE \quad (1) character-range edges by $(e.\mathit{min}, e.\mathit{max})$
\STATE \quad (2) rule-reference edges by $(h(e.\mathit{ref}))$
\STATE \quad (3) epsilon edges
\STATE Run BFS from $s_0$ using the sorted outgoing edges
\STATE Assign each state a canonical ID in discovery order

\vspace{0.5em}
\STATE {\bfseries Phase 2: Hash in the canonical order}
\STATE Let $M$ be the map from states to their canonical IDs, and let $h \gets 0$
\FOR{each state $s$ in increasing canonical ID order}
    \STATE $h \gets \mathcal{H}(h, \mathit{NODE\_TAG}, \mathbf{1}[s \in F])$
    \FOR{each edge $e$ in the sorted outgoing edges of $s$}
        \IF{$e$ is a character-range edge}
            \STATE $h \gets \mathcal{H}(h, \mathit{RANGE\_TAG}, e.\mathit{min}, e.\mathit{max}, M[e.\mathit{target}])$
        \ELSIF{$e$ is a rule-reference edge}
            \STATE $h \gets \mathcal{H}(h, \mathit{REF\_TAG}, h(e.\mathit{ref}), M[e.\mathit{target}])$
        \ELSE
            \STATE \COMMENT{$e$ is an epsilon edge}
            \STATE $h \gets \mathcal{H}(h, \mathit{EPS\_TAG}, M[e.\mathit{target}])$
        \ENDIF
    \ENDFOR
\ENDFOR

\STATE \textbf{return} $h$
\end{algorithmic}
\end{algorithm}

For the second challenge, in the cross-grammar cache, with a given rule with the FSM $A$, we will check if the token mask caches for the same FSM have been computed \autoref{fig: cgc}. If there is, then it is a cache hit. If the rules share the same lookahead assertion, then it is a perfect cache hit, and we can reuse the token mask cache directly. Otherwise, it is a partial cache hit, and we need to recheck all the uncertain tokens and the tokens that are validated by the original lookahead assertion. Then, we add the new cache to the global cache pool. In this method, most of the token mask cache will be reused. Once the size of the cross-grammar cache reaches the limit, we use LRU to evict entries. However, as we follow XGrammar’s adaptive storage method, the memory overhead of the cross-grammar cache remains low and rarely reaches this limit.

In summary, this cross-grammar cache can handle single FSMs, FSMs forming a tree reference structure, and also FSMs forming a graph with simple cycles, and maximize the cache reuse between and within grammars.

\subsection{Adaptive Token Mask Cache with Earley Parsing}
\label{earley}

Prior works, such as XGrammar~\cite{xgr}, use a token mask cache to accelerate mask generation by preprocessing the majority of tokens ahead of time. However, this design is tied to the state organization of pushdown automata. Under non-deterministic grammars, the number of PDA states can grow exponentially, which degrades both grammar compilation and runtime mask generation. To preserve the benefit of caching while improving efficiency on more complex grammars, we build a new adaptive cache mechanism on top of the Earley parser. This design inherits the cache-based acceleration strategy of prior work, while leveraging the stronger parsing efficiency of Earley parsing for complex context-free grammars.

The Earley parser~\cite{earley} maintains, at each input position, a set of partial parsing states. Each state records a production rule, a dot position within that rule, and the input position where the matching of this rule began. Together, these states define the current parsing frontier. This state organization provides a natural foundation for token-mask caching, while also requiring the cache to be defined over Earley parsing frontiers rather than the state representation used in PDA-based parsing.

Based on this observation, we design an adaptive token mask cache mechanism for the Earley parser. The key idea is to cache token validity only for the part of the parsing frontier that can directly affect the next decoding step. In Earley parsing, only scannable states, i.e., states whose next symbol is a terminal, can immediately determine whether a token may be accepted. We therefore construct caches only for these scannable states. Non-scannable states, whose next symbol is a non-terminal, are not considered in caching; instead, they will be expanded through Earley's prediction and completion operations into scannable states.

Regarding the cache content, we adapt XGrammar’s token mask categorization to the Earley parser, categorizing tokens into accepted, rejected, and context-dependent cases. The first two categories can be determined by the current partial Earley parser state, while the context-dependent tokens require the whole parsing state history to be determined. At runtime, to compute the full token mask, we first retrieve the mask cache with the current scannable states, and then check the context-dependent tokens against the full Earley context. This design reduces cache construction overhead, enables effective cache reuse, and ensures efficient mask generation for complex non-deterministic grammars.

\subsection{JIT Compilation of Adaptive Token Mask Cache}
\label{jit}
Prior efficient constrained decoding works, such as Outlines and XGrammar, have a compilation stage that computes a token mask cache for every possible state in the grammar. However, due to the intra-request dynamism in agentic tasks, one request may allow dozens or even hundreds of tools, resulting in a huge grammar that is too expensive to compile at the beginning. To avoid the large compilation overhead, we design a \textbf{configurable JIT} compilation system to amortize the grammar compilation overhead over the mask generation phase and avoid compilation for states that are never used.

To achieve JIT compilation, we design a token mask cache pool to store the generated token mask caches. This pool stores the cache corresponding to each grammar state and is initially empty. Each time we visit a new state, we will retrieve the pool for the state with the hash algorithm described in \autoref{cgc}. If cache hits, we can reuse the token mask cache directly. Otherwise, we need to generate the token mask cache at runtime and update the token mask cache pool. 

JIT compilation of the token mask cache amortizes computation from compile time to runtime. Runtime computation is overlapped with decoding, influencing per-token latency, while compilation is overlapped with prefilling and influences the time to the first token. We wish both to be hidden. It would be better hidden if we could flexibly adjust the ratio of compile-time computation amortized to runtime. Thus, we design the configurable JIT method to utilize the time. During preprocessing, we will estimate the time to generate the token mask cache for each state.  Then, we will try to calculate $K$ most time-consuming token mask, when the LLM is prefilling($K$ is a fixed value, which is adjusted for the best performance). With this method, we can overlap the time of prefilling and preprocessing, and the time of decoding and mask generation well, achieving zero-overhead token mask generation.

\subsection{Repetition State Compression}
\label{repetition}

\begin{figure}[htbp]
    \centering
    \includegraphics[width=\linewidth]{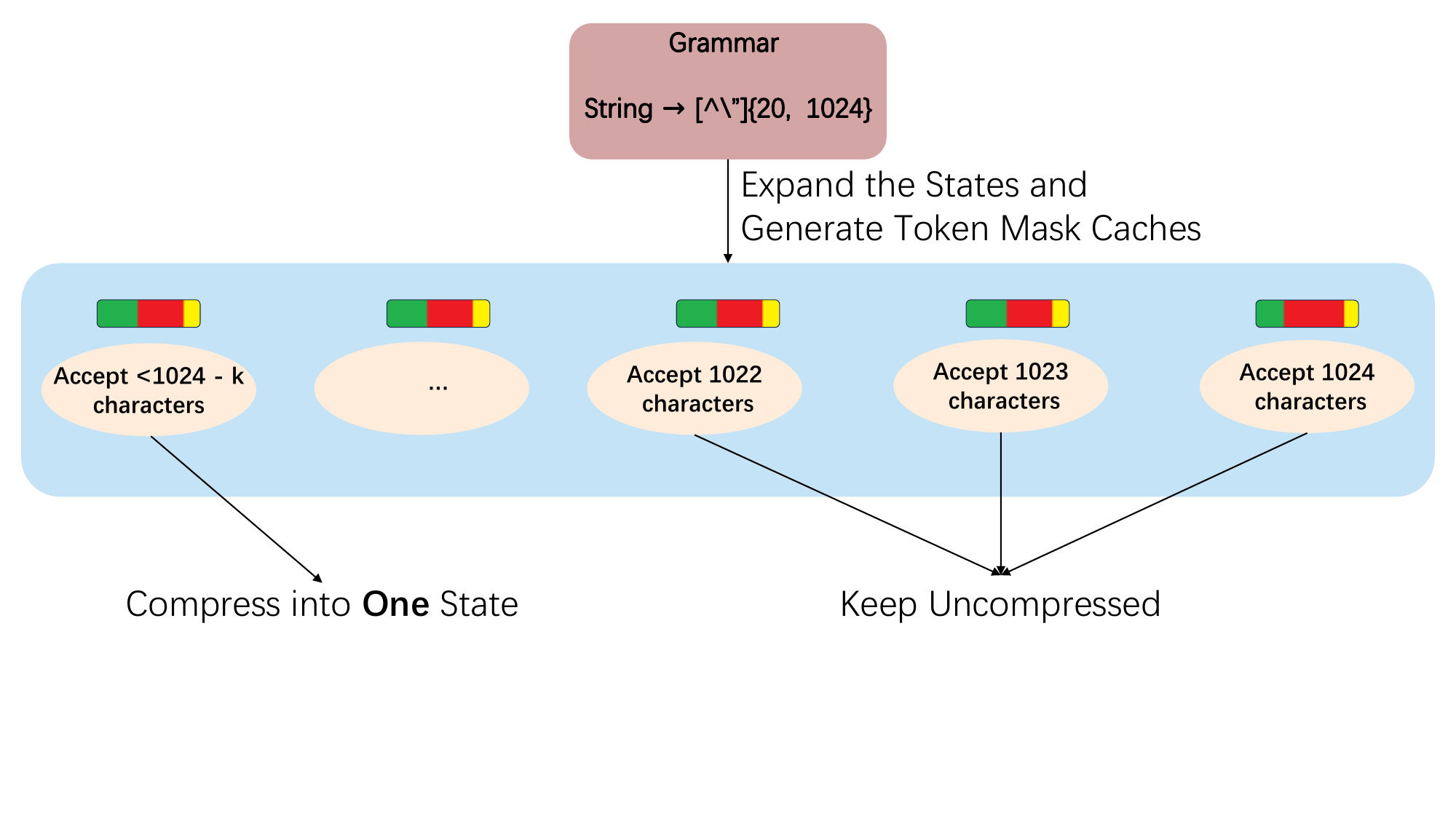}
    \caption{Repetition State Compression.}
    \label{fig: repetition}
\end{figure}

Repetition is widely used in grammar, especially in JSON schema. Keywords like \texttt{MinLength, MaxLength, MinItems, MaxItems}, etc., will generate repetition structures. If we handle the repetition structures trivially, then we need to generate a token mask cache for each possible grammar state, which is linear to the repetition times and time-consuming.

We design a repetition state compression algorithm to speed up the process. The key insight is that in many cases, the differences between states within a repetition are minimal, as illustrated in~\autoref{fig: repetition}, and we can compress the states, which bounds the size of the grammar. Formally, for a rule \texttt{R}, we introduce a special construct \texttt{R\{l, r\}} to describe the repetition structure. We require that \texttt{R} must consume at least one character to avoid repetition of zero length. The parser state for \texttt{R\{l, r\}} is \texttt{(R\{l, r\}, k)}, where k denotes the time that R has repeated. 

We can divide the raw repetition structures into three cases: (1) For \texttt{R\{l,r\}}, if \texttt{r} is small, then we expand the repetition structure as usual, since the grammar size is small. (2) If both \texttt{l} and \texttt{r} are large, then we can compress this structure. The structure will be further transformed into a sequence of \texttt{R\{l - t,r - t\}}(\texttt{t} is a chosen threshold constant) and \texttt{t} of the rule \texttt{R}. (3) If \texttt{l} is small, then we divide the \texttt{R\{l, r\}} into \texttt{R\{l,t\}} and \texttt{R\{t, r\}}. Then, we can handle each one in (1) and (2), respectively. The full algorithm is shown in Algorithm~\autoref{alg:repetition}.

After the repetition state compression algorithm, all the unexpanded repetition structures will have a subsequence of \texttt{t} times of the rule \texttt{R}. Thus, when generating token mask caches, we can perceive the repetition structures as a single state that \textbf{only} accepts sequences conforming to \texttt{R\{0, t + 1\}}, and it significantly reduces the uncertainty of the token mask caches' repetition structures. At runtime, we use the \texttt{k} of \texttt{(R\{l, r\}, k)} to check the uncertain tokens, which guarantees the correctness.

This method strikes a balance between the number of states and the uncertainty of the token mask cache. The number of states remains bounded by a constant, even for large repetition ranges, which increases the efficiency and the robustness.

\section{Evaluation}
In this section, we evaluate the efficiency and accuracy of \xg and compare \xg with state-of-the-art structured generation engines. Our experiments are motivated by the following questions:

\begin{itemize}

\item How to quantify the dynamism in agentic tasks, and how does it affect the efficiency of structured generation?  (\secref{dynamic})

\item Can \xg handle grammar compilation and mask generation efficiently? (\secref{function_calling})

\item Can \xg achieve minimal overhead for end-to-end function calling in LLM serving? (\secref{e2e})

\item How effective is each optimization technique introduced in \xg? (\secref{ablation})

\item Can \xg work correctly to constrain the LLMs' outputs in agentic tasks?
(\secref{accuracy})

\end{itemize}

For experiments focusing on the efficiency of token mask generation (\secref{dynamic}, \secref{function_calling}, \secref{ablation}, \secref{jsb}, \secref{earley_pda}), we use an AMD EPYC 9654 processor. For the end-to-end experiment (\secref{e2e}), the setup includes an Nvidia RTX 5090 GPU and an Intel(R) Xeon(R) Platinum 8470Q CPU. For accuracy evaluation (\secref{accuracy}), we utilize an Nvidia B200 GPU and an Intel(R) Xeon(R) Platinum 8570 CPU. The software versions are as follows: XGrammar, v0.1.19; llguidance, v1.2.0; Outlines, v0.2.11; and SGLang, v0.5.3.post3. All mask generation engines are run with a single thread.

\subsection{Quantifying Dynamism in Agentic Tasks}

\label{dynamic}

In this section, we quantify the dynamism in agentic tasks and justify the necessity of abstractions and optimizations introduced in this work, especially the TagDispatch intrinsic and the Cross-grammar Cache. 

\paragraph{Inter-request Dynamism.}
The main challenge for inter-request dynamism is that different requests often require different structures, making full-grammar reuse ineffective. We therefore quantify both whole-grammar overlap and reusable substructure overlap across requests.

\begin{figure}[htbp]
    \centering
    \includegraphics[width=1\linewidth]{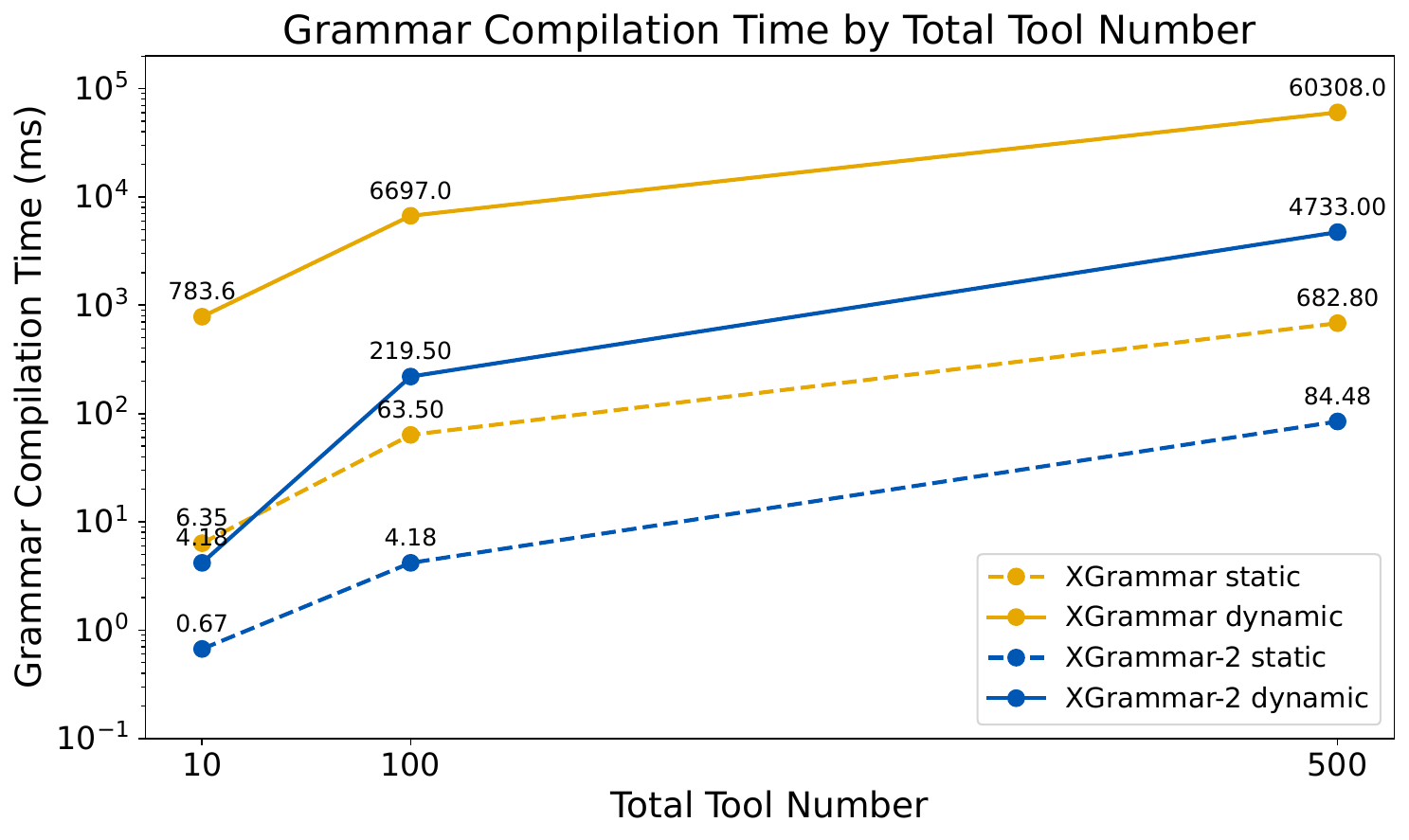}
    \caption{Average Grammar compilation time for static and dynamic workloads. Dynamic workloads significantly increase compile time.}
    \label{fig: dynamic}
\end{figure}

\begin{figure}[htbp]
    \centering
    \includegraphics[width=1\linewidth]{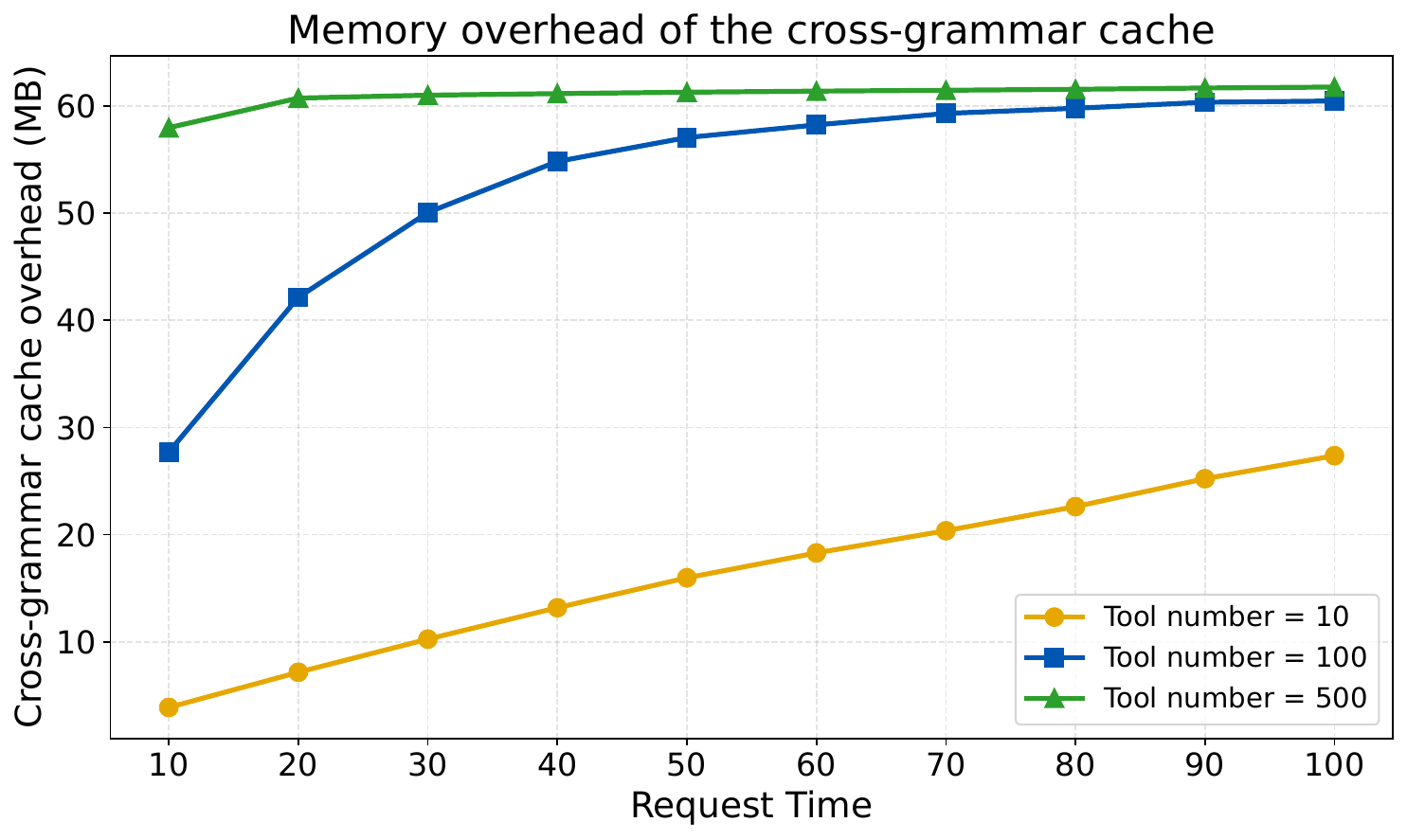}
    \caption{The memory overhead of the cross-grammar cache(MB).}
    \label{fig: cgc-memory}
\end{figure}

We choose a tool pool of 1908 distinct tools from BFCL~\cite{patil2025bfcl} and construct two scenarios, each containing 100 requests. In the \textit{static} setting, every request uses the same 10, 100, or 500 tools to build the grammar. In the \textit{dynamic} setting, each request samples 10, 100, or 500 tools uniformly at random from the tool pool. For each setting, we measure the reuse rate of full structures and substructures across requests, and report grammar compilation time in \autoref{fig: dynamic}, and the memory overhead of the cross-grammar cache is shown in~\autoref{fig: cgc-memory}.

As shown in Table~\ref{tab: dynamism}, inter-request dynamism significantly reduces the reuse of complete grammar structures in the dynamic setting. In contrast, substructure reuse remains much higher, indicating that although full grammars change frequently across requests, many underlying components can still be reused. This suggests that reuse opportunities exist primarily below the whole-grammar level.  \autoref{fig: dynamic} further shows that, in the dynamic setting, XGrammar's compilation cost increases rapidly with the number of tools due to the lack of fine-grained cache, whereas \xg scales much more gently with the cross-grammar cache. \autoref{fig: cgc-memory} also shows that the memory overhead of the cross-grammar cache will not grow rapidly as the request number grows. Due to the design of the cross-grammar cache, the memory overhead is more relevant to the total number of used tools.

Overall, inter-request dynamism makes whole-grammar reuse ineffective, since complete grammars change frequently across requests. At the same time, substantial reusable substructures remain, motivating cross-grammar reuse for efficient structured generation.

\paragraph{Intra-request Dynamism.} The main challenge for intra-request dynamism is handling free-form text together with tag-triggered dynamic structures within a single request, which is cumbersome to express in EBNF and difficult to scale.

To quantify this complexity, we consider a natural construction of plain EBNF dispatching: we first build an Aho-Corasick automaton for tag matching and then translate it into EBNF. In this translation, each automaton node corresponds to a rule, and each transition corresponds to a rule reference. We therefore record the number of automaton states, the number of automaton transitions, and the size of the resulting EBNF to reflect the amount of grammar structure needed to encode the dispatching logic. All the used tags have a common prefix like \textit{<function=}, and the rest are randomly generated.

As shown in~\autoref{tab:tagdispatch_grouped}, both the automaton size and the resulting EBNF size grow rapidly as the number of tags increases.  This indicates that implementing dispatching through plain EBNF becomes increasingly cumbersome and scales poorly. Moreover, TagDispatch is much more efficient than the plain EBNF grammar. In contrast, TagDispatch represents the dispatch structure directly, making the implementation much clearer and more compact.

\begin{table}[htbp]
\centering
\small
\setlength{\tabcolsep}{6pt}
\renewcommand{\arraystretch}{1.25}

\begin{tabular}{|c|c|c|c|c|}
\hline
\multirow{2}{*}{\shortstack{Total Tool\\Number}} 
& \multicolumn{2}{c|}{Structure Reuse Rate (\%)} 
& \multicolumn{2}{c|}{Substructure Reuse Rate (\%)} \\
\cline{2-5}
& Static & Dynamic & Static & Dynamic \\
\hline
10  & 99.0  & 0 & 99.1 & 25.2 \\
\hline
100 & 99.0 & 0 & 99.1 & 79.9 \\
\hline
500 & 99.0 & 0 & 99.1 & 95.6 \\
\hline
\end{tabular}

\caption{Structure and substructure reuse rate for static and dynamic workloads. Dynamic workloads fail to reuse the full structure, but can effectively reuse substructures.}
\label{tab: dynamism}
\end{table}

\begin{table}[htbp]
\centering
\small
\setlength{\tabcolsep}{4pt}
\renewcommand{\arraystretch}{1.15}
\begin{tabular}{c c c c c c c}
\toprule
\multirow{2}{*}{\#Tags} & \multirow{2}{*}{Total Length} 
& \multicolumn{3}{c}{Size} 
& \multicolumn{2}{c}{Compilation Time (ms)} \\
\cmidrule(lr){3-5} \cmidrule(lr){6-7}
 &  & AC $\#S$ & AC $\#E$ & EBNF 
 & EBNF & TagDispatch \\
\midrule
5   & 100   & 60  & 118   & 417  & 1008.6  & 191.4 \\
20  & 400   & 207 & 412   & 1440 & 3422.1  & 494.6 \\
50  & 1000  & 487 & 972   & 3385 & 7405.3  & 867.1 \\
100 & 2000  & 952 & 1902  & 6612 & 15483.5 & 2002.1 \\
\bottomrule
\end{tabular}
\caption{Measured sizes of the naturally constructed EBNF grammar and compilation time comparison between EBNF and TagDispatch.AC \#S means the number of states in the AC Automaton; AC \#E is the number of transitions in the AC Automaton.}
\label{tab:tagdispatch_grouped}
\end{table}

\subsection{Grammar Processing Efficiency}
\label{function_calling}

\begin{figure}[htbp]
    \centering
    \includegraphics[width=\linewidth]{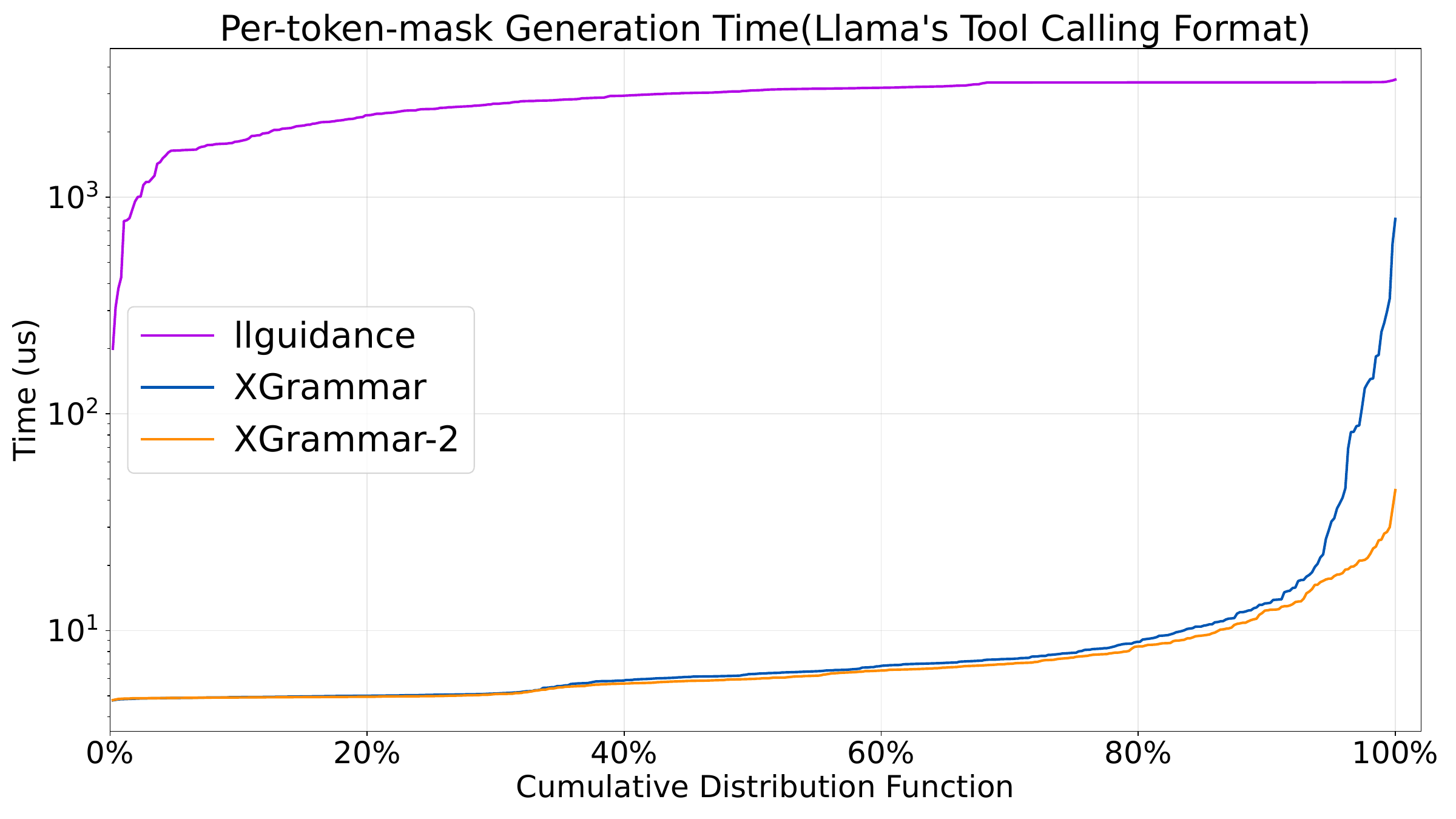}
    \includegraphics[width=\linewidth]{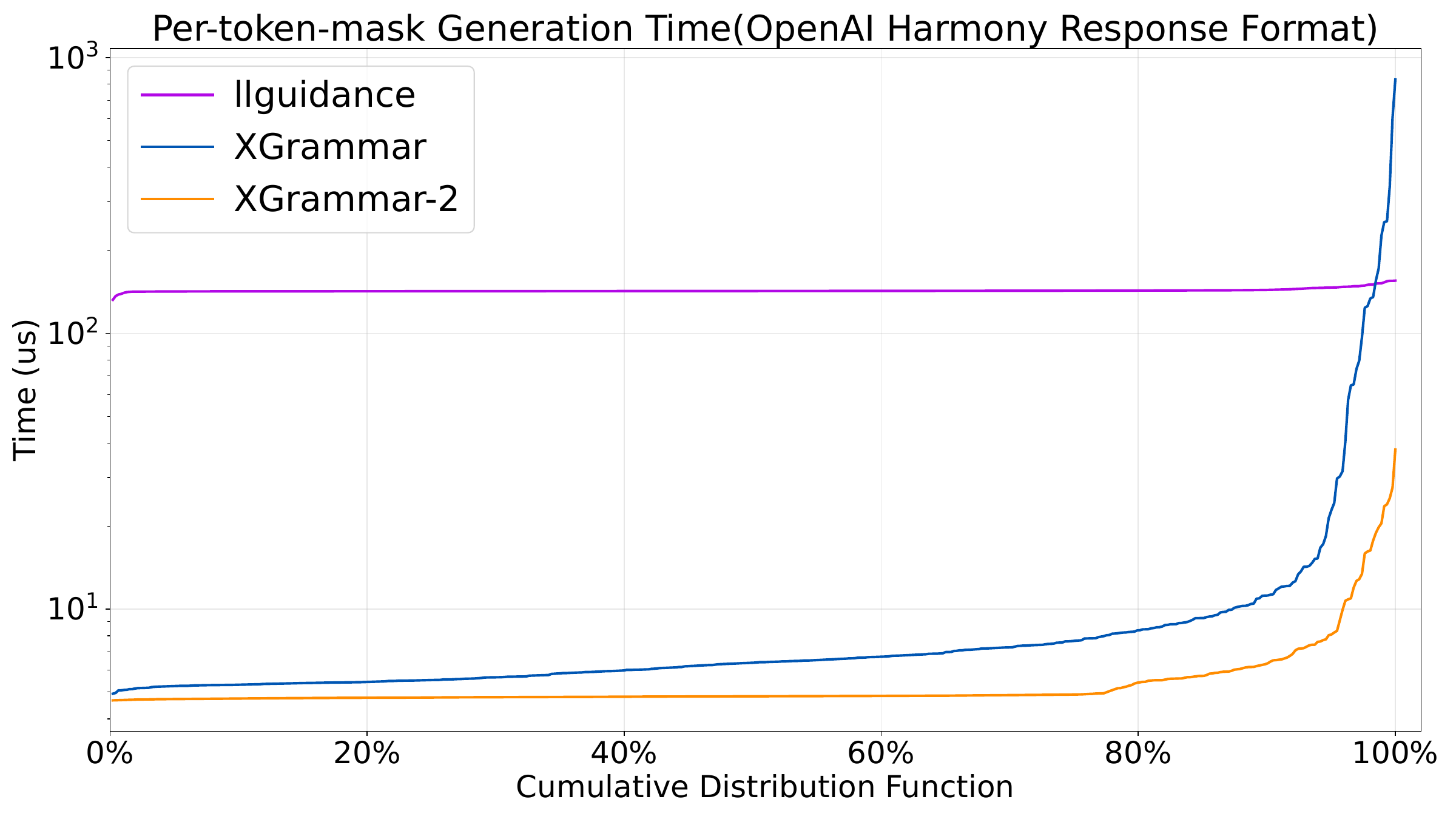}
    \caption{Average Per-token Overhead in Llama's Tool Calling Format and OpenAI Harmony Response Format.}
    \label{fig: mask_gen_per_token}
\end{figure}

\begin{figure}[htbp]
    \centering
    \includegraphics[width=\linewidth]{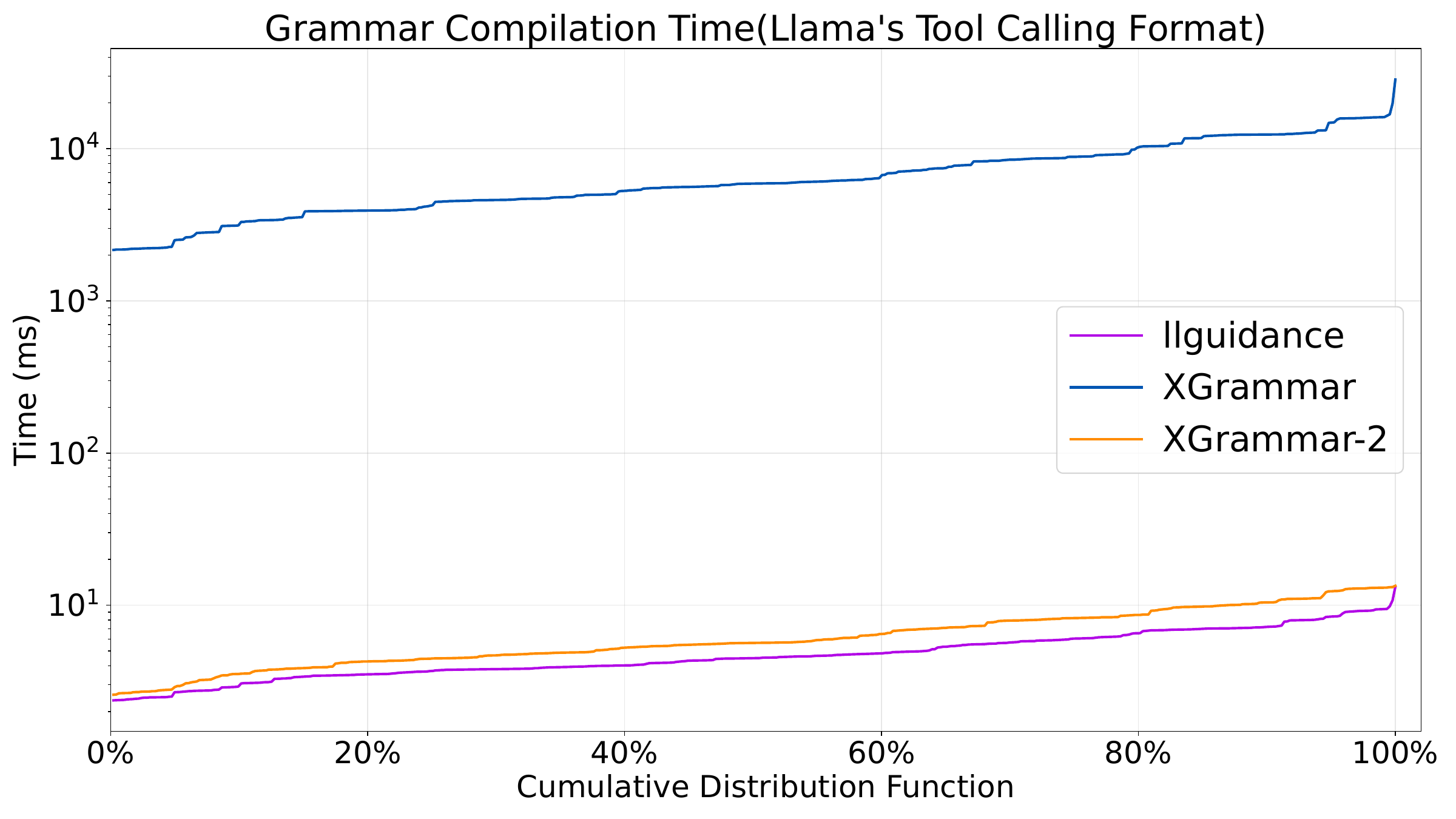}
    \includegraphics[width=\linewidth]{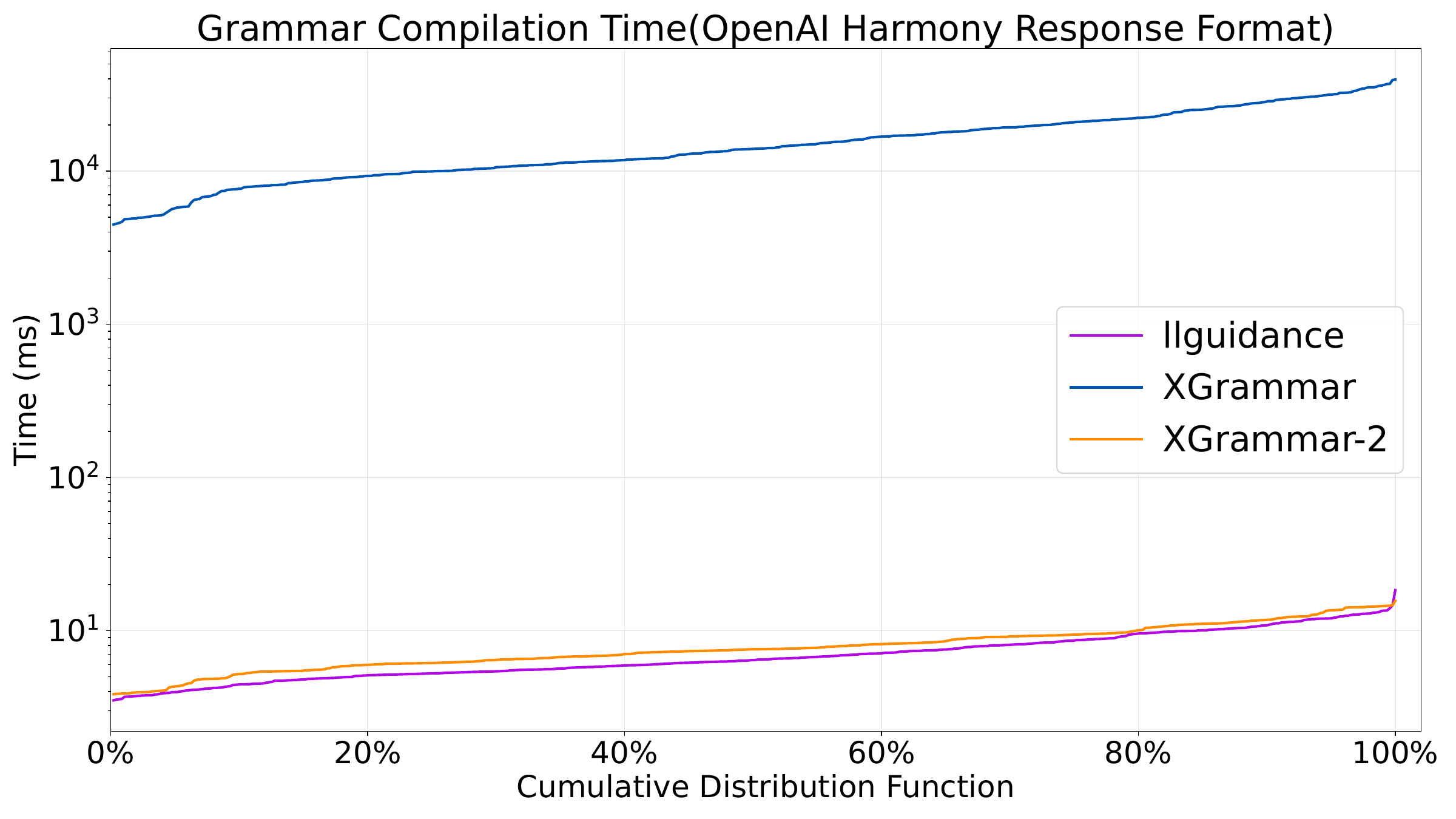}
    \caption{Compilation Time in Llama's Tool Calling Format and OpenAI Harmony Response Format.}
    \label{fig: mask_gen_compilation}
\end{figure}

In this section, we will evaluate the efficiency of grammar compilation and mask generation among several structured generation engines. We evaluate two major structures for agent tasks: function calling and response protocols are common scenarios for dynamic structured generation.

In this part, we choose CONFETTI~\cite{alkhouli2025confetticonversationalfunctioncallingevaluation} as our dataset. CONFETTI provides a collection of functions and ground-truth contexts for large language models, consisting of both natural language text and function calls. This dataset effectively simulates real-world function-calling scenarios. We modify the dataset to two common formats: Llama's tool calling format and OpenAI Harmony Response Format. The results are shown in~\autoref{fig: mask_gen_per_token},~\autoref{fig: mask_gen_compilation}. Besides, the cache hit rates of \xg are: 71.43\% (Llama's Tool Calling Format and 47.21\% (OpenAI Harmony Response Format).

The results show that \xg has an advantage in per-token overhead, while llguidance has about 250 us per-token overhead with OpenAI Harmony Response Format and a more than 1000 us per-token overhead with Llama's Tool Calling Format. XGrammar also performs well on per-token overhead. However, for dynamic structured generation tasks, mask generation engines cannot know all the grammar at the very beginning. It will introduce huge overhead if the engine needs a long compilation time. The results of compilation time show that \xg has a compilation time of about 10 ms, while XGrammar needs more than 1000 ms to compile. \xg performs well on both per-token overhead and compilation time, which demonstrates that \xg shows superior performance in grammar execution.

\subsection{End-to-end LLM Engine Evaluation}
\label{e2e}

\begin{figure}[htbp]
    \centering
    \includegraphics[width=\linewidth]{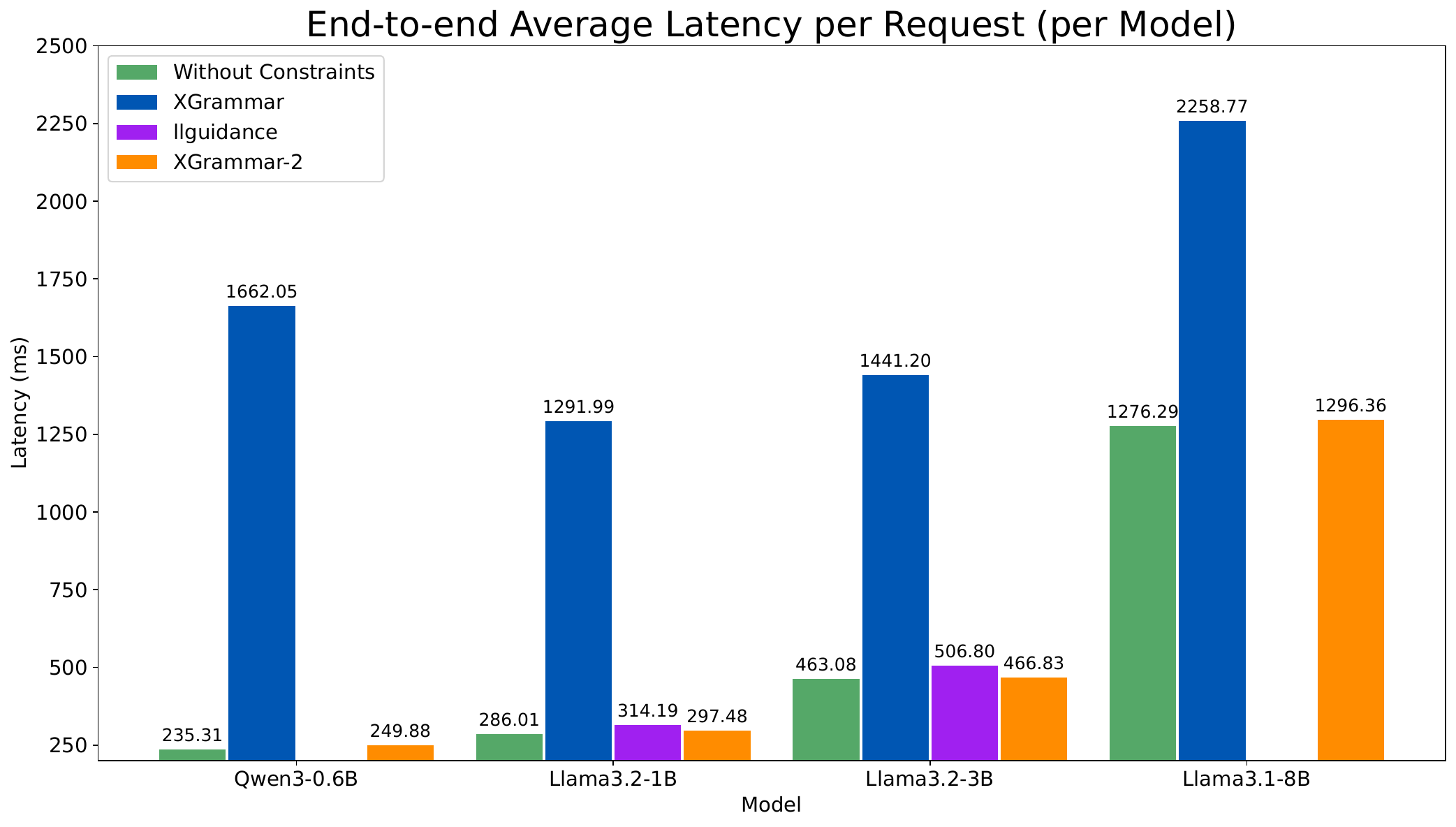}
    \caption{End-to-end Function Calling Latency.}
    \label{fig:e2e}
\end{figure}

The results in \secref{function_calling} demonstrate that \xg shows superior performance in grammar execution. In this section, we evaluate the overhead introduced by constrained decoding in real-world settings and examine whether our method achieves low-overhead structured generation for dynamic structured generation. We adopt BFCL-v3\cite{patil2025bfcl} as the dataset. BFCL-v3 is a dataset consisting of combinations of tools and prompts, which can be used to measure models' ability to call functions. Thus, we can apply structured generation engines on the models to stimulate the real serving scenarios. We use Qwen-0.6B, Llama3.2-1B, Llama3.2-3B-Instruct, and Llama3.1-8B as the test models, and run the test with SGLang. SgLang-v0.5.3.post3 with Outlines-v0.2.11 cannot support dynamic structured generation like tool-calling. SgLang-v0.5.3.post3 with llguidance-v1.2.0 can support dynamic structured generation, but it results in empty outputs for Qwen3-0.6B and induces language drift from pure English to other languages in Llama3.1-8B. The results are shown in~\autoref{fig:e2e} and ~\autoref{tab: throughput}.

\begin{table}[htbp]
\centering
\begin{tabular}{|c|c|c|c|c|}
\hline
\textbf{Model Name} &
\textbf{Type} &
\multicolumn{3}{c|}{\textbf{Batch Size}} \\
\cline{3-5}
 &  &
\textbf{1} &
\textbf{16} &
\textbf{128} \\
\hline

\multirow{2}{*}{Qwen3-0.6B}
 & XGrammar   & 462 & 1712 & 3021 \\
 & \xg & 599 & 4287 & 9475 \\
\hline

\multirow{2}{*}{Llama-3.2-1B}
 & XGrammar   & 274 & 861 & 1147 \\
 & \xg & 441 & 2933 & 6640 \\
\hline

\multirow{2}{*}{Llama-3.2-3B}
 & XGrammar   & 139 & 597 & 791 \\
 & \xg & 184 & 1655 & 3830 \\
\hline

\multirow{2}{*}{Llama-3.1-8B}
 & XGrammar   & 83 & 525 & 738 \\
 & \xg & 96 & 920 & 1938 \\
\hline

\end{tabular}
\caption{The output token throughput (token/s) With Different Models and Batch Size.}
\label{tab: throughput}
\end{table}

The results in~\autoref{fig:e2e} show that compared to XGrammar, \xg has about a 7x speedup over the end-to-end latency, and also a larger total token throughput. Besides, the gap between the result of \xg and the result without constraints is no more than 6\%. Compared with llguidance, \xg shows a small latency and better compatibility. The output token throughput in~\autoref{tab: throughput} also shows that \xg is superior to XGrammar. This demonstrates that \xg can support dynamic structured generation efficiently.

\subsection{Ablation Study of Optimization Techniques}
\label{ablation}

In this section, we further investigate the efficiency improvements brought by our various optimizations to better illustrate the reasons for our design decisions. We start with a baseline implementation using the Earley parser and without any of the optimizations. Based on the baseline, we incrementally apply the proposed optimizations, namely JIT compilation, cross-grammar cache, and repetition state compression. We choose JSONSchemaBench \cite{jsonschemabench} as the dataset. JSONSchemaBench collects about 11k JSON Schemas from about 20 lines to more than 200k lines. This dataset can be used to measure each optimization technique from multiple angles.

\begin{table}[htbp]
\centering

\caption{Ablation study of optimization techniques.}

\label{tab: ablation}

\begin{tabular}{ccc}

\hline
\textbf{Optimization} & \textbf{preprocessing} & \textbf{time to generate} \\
& \textbf{time($ms$)} & \textbf{ the mask($\mu s$)} \\
\hline
Baseline & 4960.04 & 45.50 \\
$\downarrow$ & & \\

+JIT & 612.07 & 722.47 \\
$\downarrow$ & (8.1×↓) & (15.9×↑) \\

+ Cross-grammar & 534.80 & 333.75 \\
Cache & (1.1×↓) & (2.2×↓) \\
$\downarrow$ & & \\

+Repetition State & 5.37 & 126.49 \\ 
Compression & (99.6×↓) & (2.6×↓) \\

\hline
\end{tabular}

\end{table}

The results show that JIT serves as a general optimization technique that substantially improves preprocessing time, although it introduces additional overhead to generate the mask.  Cross-Grammar Caching can generally reduce the time to generate the mask to an acceptable level, and keep the mask generation time low in cache-hit cases. Besides, Repetition Compression achieves significant improvements on some long-tail cases because it can ensure a constant process time on repetition structures.  We also evaluate the benefit of the Earley Parser, and the result is in~\autoref{earley_pda}.

\section{Related Work}
Several works focus on LLMs' structured generation. In the very beginning, ~\cite{yin2017syntacticneuralmodelgeneralpurpose} proposed a new architecture to guide the output of models with pre-defined rules. PICARD\cite{scholak2021picardparsingincrementallyconstrained} designs an algorithm to parse incrementally for Constrained Auto-Regressive decoding from language models. \cite{mudgal2024controlleddecodinglanguagemodels} proposes controlled decoding for alignment of LLMs. \cite{wang2023grammarpromptingdomainspecificlanguage} explores utilizing prompts to specify the LLMs' generation structure.  \cite{ rozière2024codellamaopenfoundation, codealpaca, li2023starcodersourceyou} design finetuning technologies for higher quality structured generation. \xg is orthogonal to these methods, and can be easily combined with them to better support structured generation.
    
Several frameworks have been proposed to support constrained decoding.
Outlines~\cite{willard2023efficientguidedgenerationlarge} designs an FSM-based lexer and parser, and it caches several of the most common lexer tokens to speed up. However, when the LLMs output contains multiple lexemes, the caching algorithm cannot perform well.
XGrammar~\cite{xgr} utilizes pushdown automata as the parsing backend, and it caches all the token mask caches in advance for better performance at runtime. However, it will suffer from a long compilation time in dynamic structured generation.
llguidance~\cite{githubGitHubllguidance} employs an Earley parser to parse the prior LLM output, and it applies a series of optimization algorithms to reduce per-token latency. But it targets specific JSON structures and has not yet generalized well to dynamic structured generation in agentic tool-calling use cases. WGRAMMAR~\cite{wang2025wgrammarleveragepriorknowledge} provides a structural template to reuse the token mask caches in the template to accelerate. But it has not generalized it to all similar grammar structures. \xg builds on top and complements these previous approaches by enabling dynamic structured generation through tag dispatch, JIT-based cross-grammar cache mechanism, Earley parser, and the token mask cache.

Several LLM serving engines~\cite{mlc-llm,zheng2024sglangefficientexecutionstructured,kwon2023efficient,hiworldwzj2024ModelTCLightLLM} employ different techniques to support efficient LLM generation for multiple concurrent users. They design various techniques such as continuous batching~\cite{orca} for dynamic request scheduling, low-level KV cache technique PagedKVCache~\cite{kwon2023efficient} for efficient memory management, and ~\cite{ye2025flashinferefficientcustomizableattention} for a more customizable and efficient attention engine. These LLM serving engines can leverage \xg for more efficient dynamic structured generation.

\section{Conclusion}

We proposed \xg, an efficient structured generation engine for LLMs' dynamic structured generation tasks. We designed a dynamic dispatching semantics to efficiently support dynamic structured generation. Additionally, we designed a cross-grammar caching mechanism based on the Earley parser. We also introduce just-in-time (JIT) compilation for token mask caching, building upon the work of XGrammar. Finally, we design a repetition compression algorithm to handle several long-tail cases. Experimental results demonstrate that \xg supports dynamic structured generation tasks with near-zero overhead. We hope that \xg can significantly enhance the efficiency of dynamic structured generation tasks.

\begin{acks}

This work is supported in part by Bosch and gifts from NVIDIA and Google. We also acknowledge the support of DGX B200 from NVIDIA.
We would also like to thank, listed alphabetically, Databricks, the SGLang team, the TensorRT-LLM team, the vLLM team, and xAI, as well as Yi Wang, Xinyu Yang, Jieyu Zhang, Wenxin Zheng, and Ligeng Zhu, for their insightful feedback.

\end{acks}

\bibliographystyle{ACM-Reference-Format}
\bibliography{references}

\appendix
\section{The Hash Algorithm for Simple Cycle Structure}

Algorithm~\autoref{alg:fsm_hash} presents the procedure for hashing FSMs in a simple cycle structure. In this setting, all FSMs referenced by those in the cycle are first hashed using Algorithm~\autoref{alg:fsm_hash}. Consequently, for each FSM in the cycle, exactly one referenced FSM remains unhashed, namely the next FSM in the cycle. We therefore assign a shared placeholder constant $X$ to these unresolved references and compute a hash for each FSM using Algorithm~\autoref{alg:fsm_hash}. This yields a \textbf{local} hash value for each FSM, which captures only the individual FSM but not the overall cycle structure. Finally, we combine the local hash values of all FSMs in the cycle to derive the final hash for each FSM. Since the hash function is non-commutative, the resulting final hash values are unique.

\begin{algorithm}[htbp]
\caption{Handle Simple Cycle Structure in FSM Reference}
\label{alg: cycle}
\begin{algorithmic}
\STATE {\bfseries Input:} a series of local hash values of simple-cycle FSMs $L_0,L_1,...,L_n$
\STATE {\bfseries Output:} a series of final hash values of simple-cycle FSMs $H_0,H_1,...,H_n$
\vspace{0.75em}

\FOR{$i\text{ in range}(n+1)$}
    \STATE $H_i \gets 0$
    \FOR{$j\text{ in range}(n+1)$}
        \STATE $H_i \gets \mathcal{H}(H_i,\, L_{[(i + j) \bmod |L|]})$
    \ENDFOR
\ENDFOR

\STATE \textbf{return} $H_0,H_1,...,H_n$

\end{algorithmic}
\end{algorithm}

\section{The Algorithm for Repetition State Compression}

Algorithm~\autoref{alg:repetition}
shows the algorithm to perform the repetition state compression algorithm in detail.

\begin{algorithm}
\caption{Repetition State Compression Algorithm}
\begin{algorithmic}
\label{alg:repetition}
\STATE {\bfseries Input:} A triplet $(min,\, max,\, context)$
\STATE {\bfseries Output:} A expression $expr$
\STATE {\bfseries Const:} $kRepetitionThreshold \gets t$

\vspace{0.75em}

\IF{$max \leq t$}
    \STATE $expr \gets \textsc{Expand}(min,\, max,\, context)$
    \STATE \textbf{return}
\ENDIF

\IF{$min < t$}
    \STATE $other\_choices \gets \textsc{Expand}(min,\, t,\, context)$
    \STATE $choice \gets \textsc{Concat}(\textsc{Repeat}(t,\, max,\, context), $\\$
    \textsc{Expand}(0,\, max - t,\, context))$
    \STATE $expr \gets \textsc{Union}(choice,\, other\_choices)$
    \STATE \textbf{return}
\ENDIF

\FOR{$i \in \text{range}(t)$}
    \STATE $expr \gets \textsc{Concat}(expr,\ context)$
\ENDFOR

\STATE  $expr \gets \textsc{Concat}(expr,\ \textsc{Repeat}(min - t,\, max - t,\, context))$

\vspace{0.75em}

\FUNCTION{\textsc{Repeat}($min, max, context$)}
    \STATE \textbf{return} a repetition expression that accepts $context$ at least $min$ times and at most $max$ times
\ENDFUNCTION

\vspace{0.75em}

\FUNCTION{\textsc{Expand}($min, max, context$)}
    \STATE \textbf{return} an explicit expansion equivalent to the repetition expression
\ENDFUNCTION

\vspace{0.75em}

\FUNCTION{\textsc{Union}($expr_1, expr_2$)}
    \STATE \textbf{return} an expression that matches either $expr_1$ or $expr_2$
\ENDFUNCTION

\vspace{0.75em}

\FUNCTION{\textsc{Concat}($expr_1, expr_2$)}
    \STATE \textbf{return} an expression that matches $expr_1$ followed by $expr_2$
\ENDFUNCTION

\end{algorithmic}
\end{algorithm}

\section{More Explanation of the Hash Algorithm}

For most FSMs, this algorithm generates a consistent hash value. However, there are two cases where it may produce different hash values for FSMs with the same structure: (1) the FSM is not a deterministic finite automaton (DFA); (2) there are duplicated FSMs in the grammars, and they are referenced by a common FSM. In these cases, the algorithm may generate inconsistent hash values. Nevertheless, this does not undermine the sufficiency of the algorithm: if two FSMs share the same hash value, they must have the same structure. In addition, in our implementation, we attempt to transform most FSMs into DFAs. Moreover, since we have a deterministic conversion function for JSON Schemas and regular expressions, two FSMs with the same structure are likely to produce the same hash value due to this deterministic transformation. As a result, we can detect and reuse identical structures within and across grammars maximally.

\section{Discussion on the Parameter K in Configurable JIT}

The parameter $K$ depends on both the GPU and the CPU. Tuning it with elaboration can improve the efficiency and stability.
We swept $K$ under the setup described in \autoref{function_calling}, using Llama's tool-calling format.
The results are summarized in ~\autoref{tab:k-sweep}.

\begin{table}[htbp]
  \centering
  \begin{tabular}{@{}ccccc@{}}
    \toprule
    $K$ & Compilation & Avg.\ TPOM & Max TPOM & P99 TPOM\\
    \midrule
     0 & 14.45 ms & 12.76 $\mu$s & 76.08 $\mu$s & 48.15 $\mu$s \\
     5 & 18.24 ms & 12.80 $\mu$s & 74.78 $\mu$s & 44.59 $\mu$s \\
    10 & 20.07 ms & 12.68 $\mu$s & 67.80 $\mu$s & 42.49 $\mu$s \\
    \bottomrule
  \end{tabular}
  \caption{Effect of $K$ on compilation time and TPOM metrics.}
  \label{tab:k-sweep}
\end{table}

Across this sweep, average TPOM is almost unchanged while compilation time increases with~$K$.
P99 and max TPOM decrease from $K{=}0$ to $K{=}10$, indicating a trade-off between compilation cost and tail latency.

\section{XGrammar's Adaptive Token Mask cache Generation Algorithm}

In XGrammar, all grammars are processed as a group of FSMs. During compilation, for each state of the FSMs, a corresponding adaptive token mask cache is generated. Each adaptive token mask cache consists of three parts:

\begin{itemize}
    \item Accepted tokens: tokens that can be accepted by the FSM and thus conform to the grammar.
    \item Rejected tokens: tokens that will be rejected by the FSM and therefore do not conform to the grammar.
    \item Uncertain tokens: tokens that can reach the final state(s) of the FSMs without consuming all their characters. The remaining part must be checked at runtime.
\end{itemize}

At runtime, we collect all the current states. Tokens that can be accepted by at least one adaptive token mask cache are directly marked as accepted. For the remaining tokens, if a token is marked as uncertain in at least one adaptive token mask cache, we further check whether it can be accepted given the current states. If so, it is also marked as accepted. All other tokens are marked as rejected. Through this process, a final token mask is generated.

\section{Earley's Parsing Algorithm}

The efficiency of the Earley parser comes from its well-designed algorithm, which applies dynamic programming. During parsing, it records the current state (the rule and the position within the rule), the number of characters consumed, and the starting position of the current rule. Based on the information, the parser performs three basic operations: \textit{predict}, \textit{scan}, and \textit{complete}. \textit{Predict} applies when the current position in a rule references another rule; in this case, the parser transitions to the referenced rule and applies Earley’s algorithm recursively. \textit{Scan} applies when the rule expects a character, and the parser checks whether the current character can be accepted by the state. \textit{Complete} applies when a rule reaches its end; the parser then returns to its parent states (which may be multiple) and advances them. With these three operations, the Earley parser efficiently exploits common substructures among different rules, thereby improving parsing performance.

\section{Mask Generation Efficiency on JSON Schemas}
\label{jsb}

\begin{figure}[htbp]
    \centering
    \includegraphics[width=\linewidth]{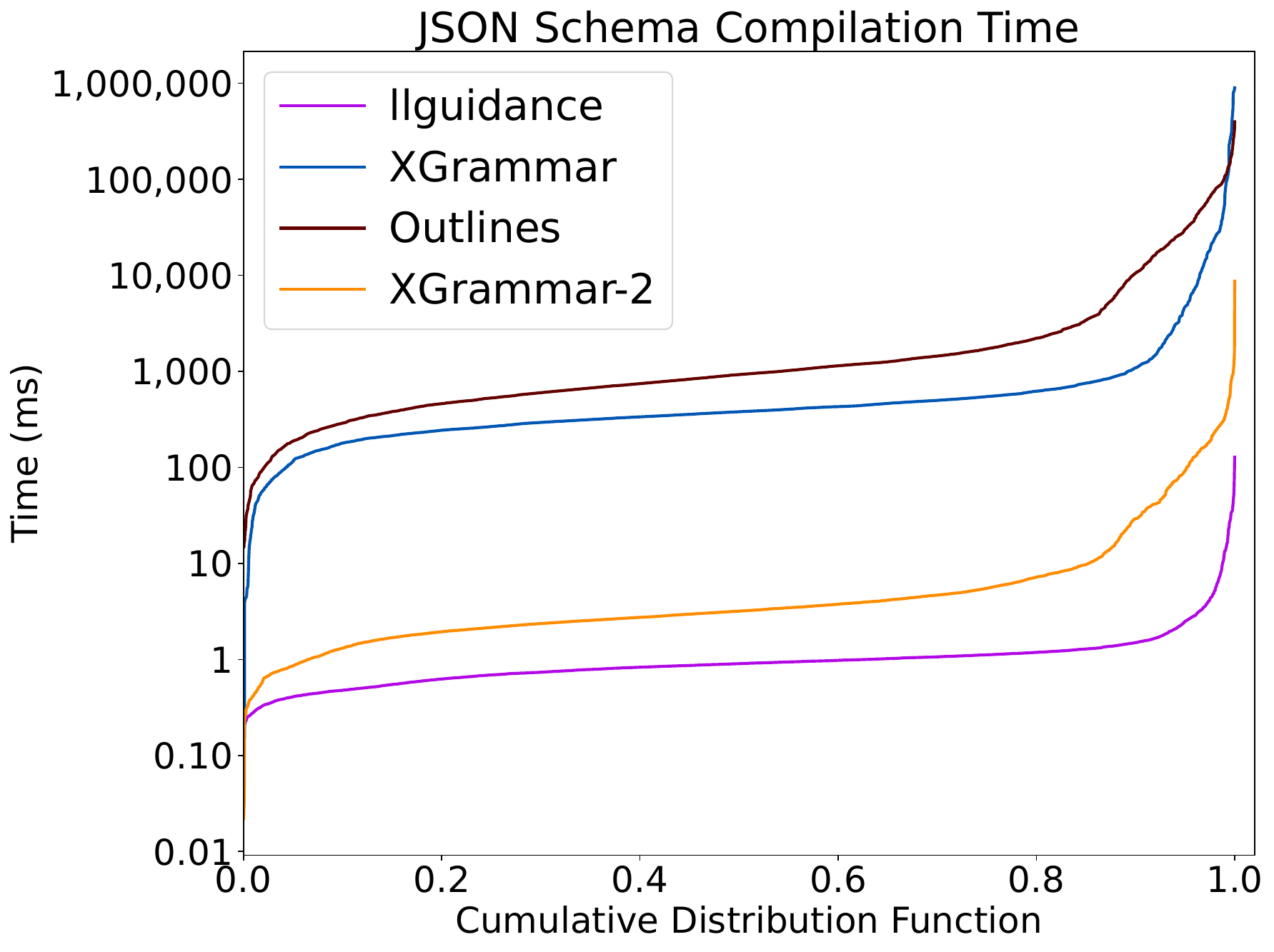}
    \includegraphics[width=\linewidth]{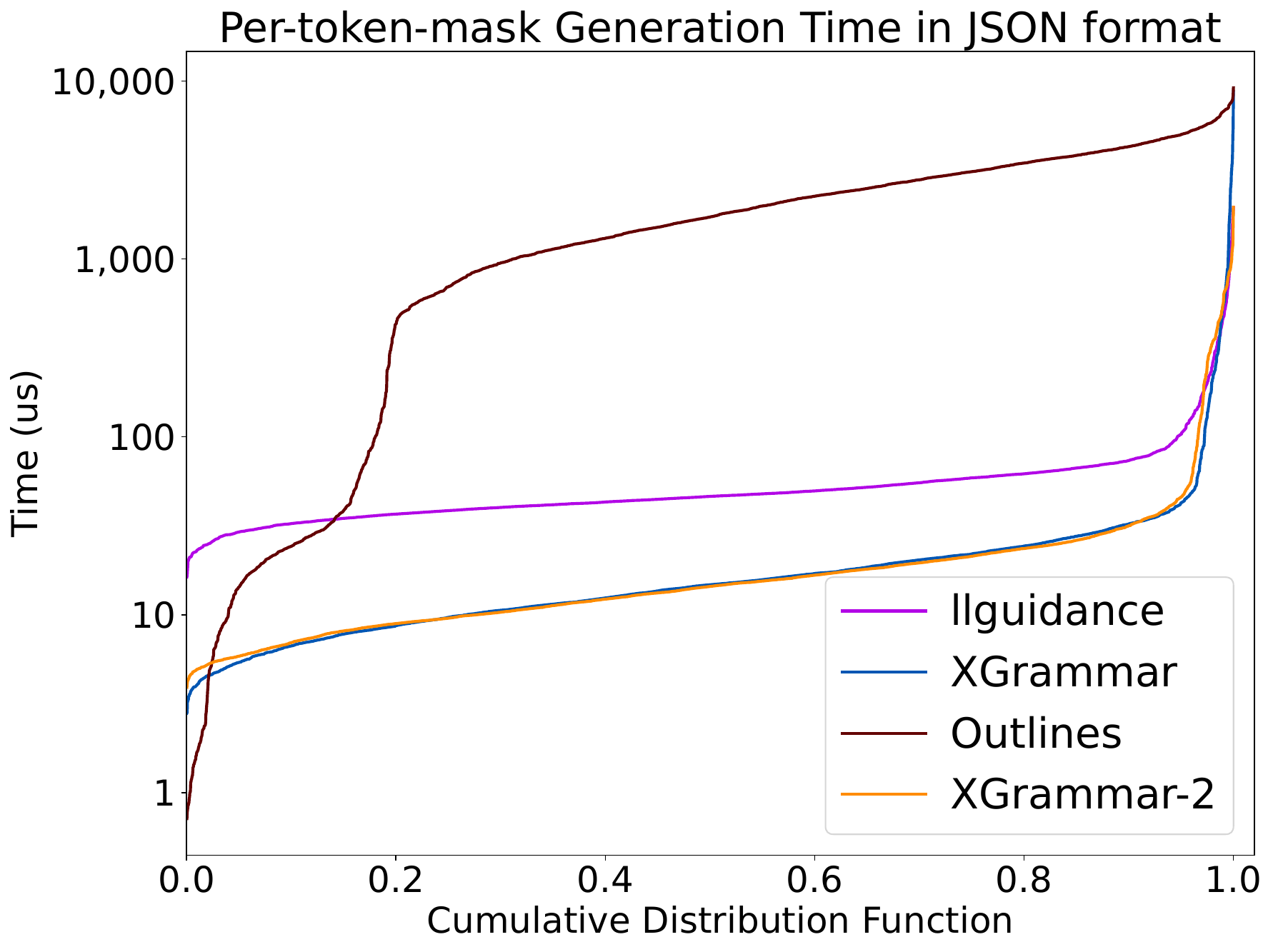}
    \caption{JSONSchemaBench.}
    \label{fig: jsonschema_bench}
\end{figure}

Although this paper focuses on dynamic structure generation in agentic use cases, it is still interesting to see how \xg performs on generations with pre-defined static JSON schemas. The dataset in JSONSchemaBench~\cite{jsonschemabench}. The results are in \autoref{fig: jsonschema_bench}.
\xg can also perform well on static structured generation tasks. Additionally, \xg brings improved grammar compilation time to compile most JSON Schemas within 1 ms.

\section{Ablation Study Between the Earley Parser and PDA Based Parser}
\label{earley_pda}

\begin{figure}[htbp]
    \centering
    \includegraphics[width=\linewidth]{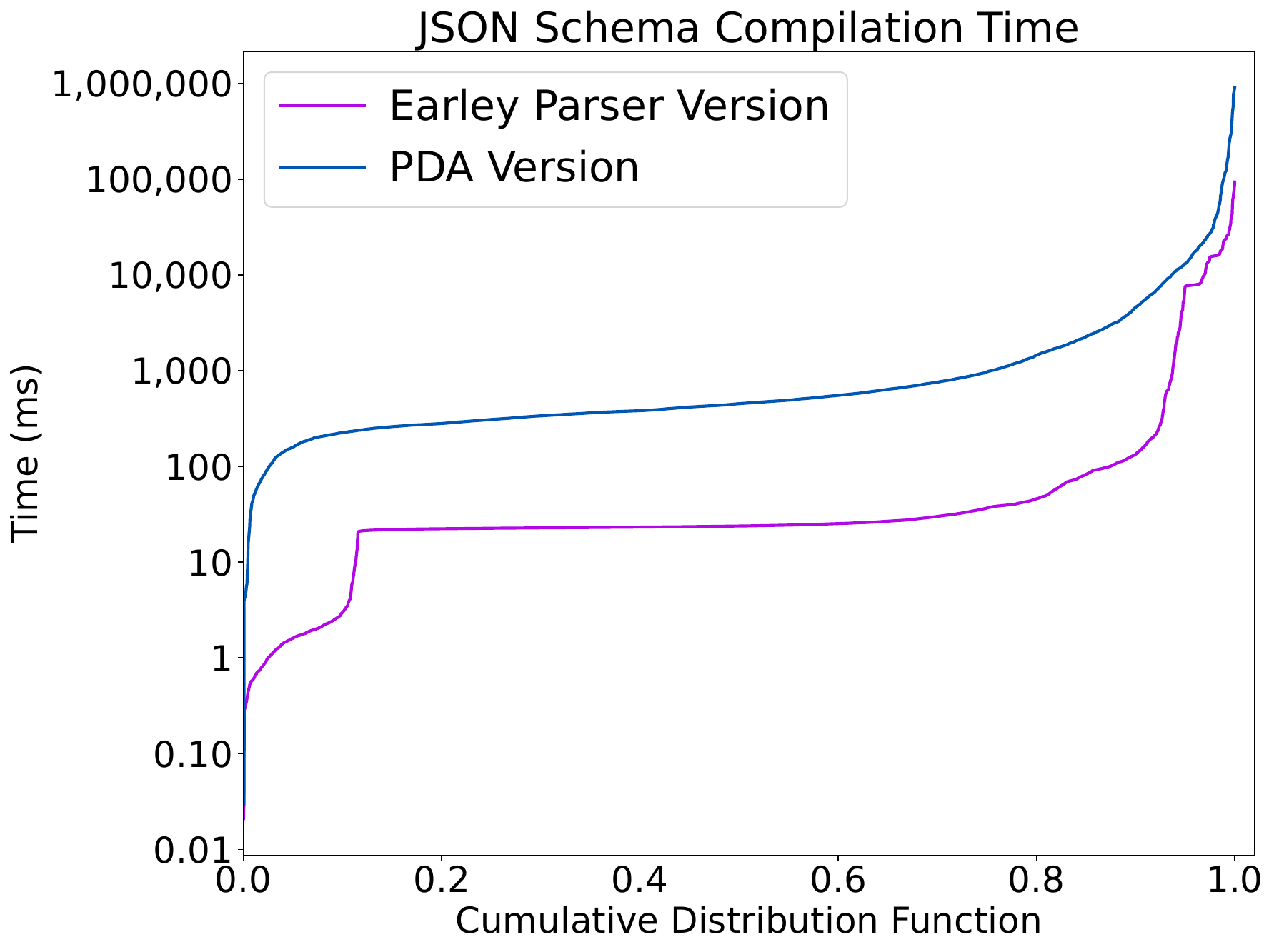}
    \caption{Comparison between the Earley Parser and PDA on JSONSchemaBench.}
    \label{fig: earley_pda}
\end{figure}

We also want to measure the advantages of the Earley Parser as an ablation study. Thus, we evaluate the efficiency of \xg, with PDA based parser and the Earley Parser, respectively, and both of them will compile the JSON schemas ahead of time. The dataset is JSONSchemaBench~\cite{jsonschemabench}, and the result in \autoref{fig: earley_pda}
shows that the Earley Parser can significantly reduce the grammar compilation. Note that the long-tail is caused by the huge inputs, instead of the complexity of the algorithm.

\section{Correctness and Task-level Effectiveness}
\label{accuracy}

By construction, constrained decoding guarantees that generated outputs conform to the target structure (e.g., JSON schema or tool-calling format). \xg preserves the same constraint semantics as XGrammar, and thus both achieve 100\% schema-valid tool-call arguments whenever a tool call is produced; the difference is efficiency (Section~\ref{e2e}).

\begin{table}[htbp]
\centering
\begin{tabular}{|c|c|c|c|}
\hline
\textbf{Model Name} &
\textbf{Type} &
\makecell{\textbf{Correct}\\\textbf{Call Rate}} &
\makecell{\textbf{Correct}\\\textbf{Schema Rate}} \\
\hline

\multirow{2}{*}{Llama-3.2-1B}
 & w/o \xg & 6.07\% & 22.07\% \\
 & w/ \xg & 32.84\% & 100.00\% \\
\hline

\multirow{2}{*}{Llama-3.2-3B}
 & w/o \xg & 33.12\% & 40.70\% \\
 & w/ \xg & 77.75\% & 100.00\% \\
\hline

\multirow{2}{*}{Llama-3.1-8B}
 & w/o \xg & 59.48\% & 66.95\% \\
 & w/ \xg & 80.93\% & 100.00\% \\
\hline

\multirow{2}{*}{Llama-3.1-70B}
 & w/o \xg & 45.60\% & 51.94\% \\
 & w/ \xg & 86.41\% & 100.00\% \\
\hline

\end{tabular}
\caption{The function calling accuracy rate and the JSON schema validity rate.}
\label{tab: accuracy}
\end{table}

To validate end-to-end correctness and quantify task-level impact in realistic agent settings, we evaluate on BFCL-v3~\cite{patil2025bfcl}. As shown in Table~\ref{tab: accuracy}, grammar-constrained decoding (\xg) substantially improves BFCL function-calling outcomes for most models, primarily by eliminating malformed tool calls (e.g., invalid JSON or schema violations) that would otherwise be unexecutable and scored as failures. Constraint enforcement can also narrow the gap between small and large models; for example, \xg enables Llama-3.2-3B to outperform an unconstrained Llama-3.1-70B baseline on BFCL.

\section{Formal Definitions of the Earley Parser and the Token Mask Generation with Cache}

Table 7 shows the formal definition of the Earley Parser~\cite{opedal2023efficientsemiringweightedearleyparsing}, and the formal definition of the token mask generation with cache. In the Table 7, Grammar Production represents a series of rules in the format of $\text{rule}\to \gamma$, where $\gamma(\mu,\  \rho)$ is the sequence of the rule. $A, B$ represents the non-terminal elements in the sequence, and $a$ represents the terminal element. $\mathcal{V}$ is the vocabulary of the tokenizer. For a token mask cache, $\mathcal{A}$ means the set of accepted tokens, $\mathcal{U}$ means the set of uncertain tokens, and $\mathcal{R}$ means the set of rejected tokens.

\label{formal}

\begin{figure*}[htbp]
    \centering
    \includegraphics[width=\linewidth]{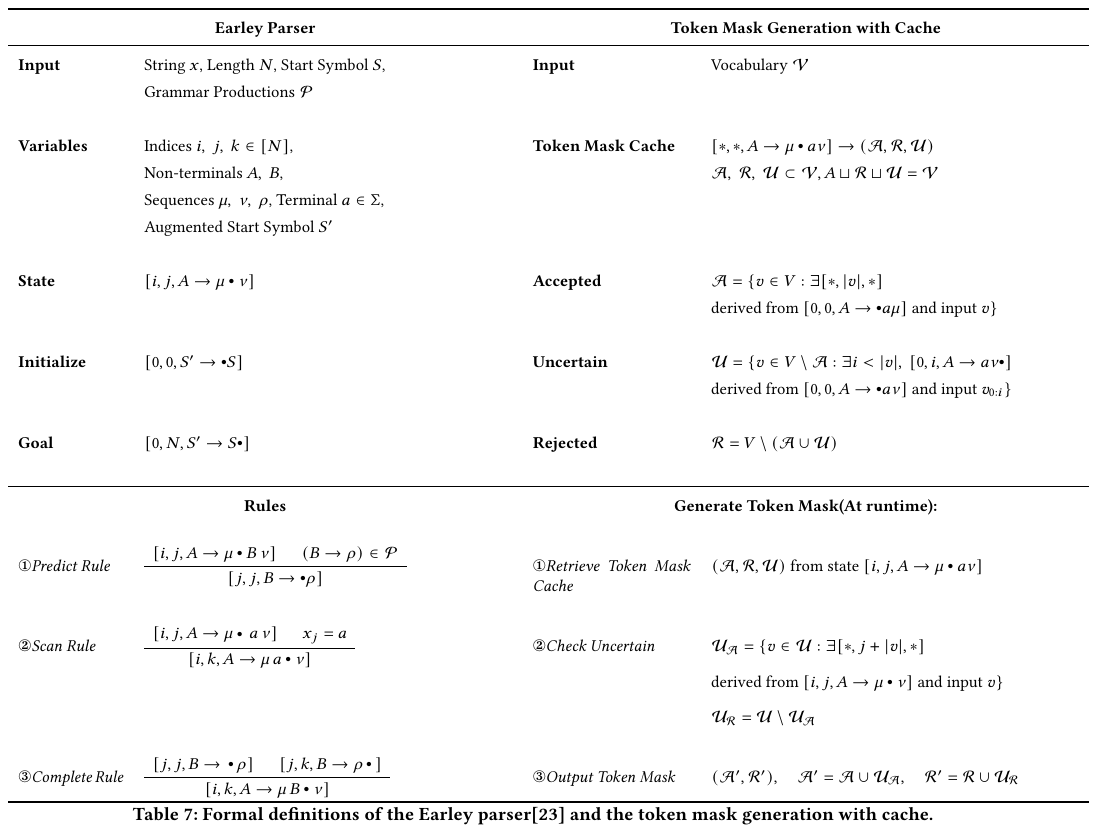}
\end{figure*}

\end{document}